\documentclass[runningheads]{llncs}

\usepackage[mobile]{eccv}
\usepackage{eccvabbrv}
\usepackage{graphicx}
\usepackage{booktabs}
\usepackage{mathtools}
\let\citep\cite
\let\citet\cite
\usepackage{multirow}
\usepackage{algorithm}
\usepackage{algpseudocode}
\usepackage[table]{xcolor}
\usepackage{adjustbox}
\usepackage{placeins}
\usepackage{tabularx,colortbl,array}
\definecolor{lightpink}{rgb}{0.980, 0.941, 0.941}
\usepackage[accsupp]{axessibility}
\usepackage{hyperref}

\usepackage{amsmath,amsfonts,bm}

\newcommand{\norm}[1]{\left\Vert #1 \right\Vert}

\def\eqref#1{equation~\ref{#1}}

\def\1{\bm{1}}

\def\vm{{\bm{m}}}

\def\vv{{\bm{v}}}

\def\vx{{\bm{x}}}

\def\mI{{\bm{I}}}

\def\mX{{\bm{X}}}

\def\mZ{{\bm{Z}}}

\DeclareMathAlphabet{\mathsfit}{\encodingdefault}{\sfdefault}{m}{sl}
\SetMathAlphabet{\mathsfit}{bold}{\encodingdefault}{\sfdefault}{bx}{n}

\newcommand{\E}{\mathbb{E}}
\newcommand{\Ls}{\mathcal{L}}

\begin{document}

\title{Momentum Guidance: Plug-and-Play Guidance for Flow Models}

\newcommand{\samethanks}[1][\value{footnote}]{\footnotemark[#1]}

\author{Runlong Liao\inst{1}\thanks{Co-first authors.} \and
Jian Yu\inst{1}\samethanks \and
Baiyu Su\inst{1} \and\\
Chi Zhang\inst{1} \and
Lizhang Chen\inst{1} \and
Qiang Liu\inst{1}}

\authorrunning{}

\institute{The University of Texas at Austin, Austin, TX 78712, USA\\
\email{liaorl@cs.utexas.edu}}

\maketitle
\pagestyle{titlepage}

\begin{abstract}
Flow-based generative methods offer a simple and effective framework for high-fidelity generation, yet pretrained flow models are rarely used in their vanilla conditional form: in image generation, samples without guidance often appear diffuse and lack fine-grained detail. Existing guidance techniques such as classifier-free guidance (CFG) improve fidelity but reduce sample diversity. We introduce \textbf{Momentum Guidance (MG)}, a guidance method that improves sample quality by extrapolating the current velocity away from an exponential moving average of past velocities along the ODE trajectory, while preserving the standard one-evaluation-per-step cost. MG provides gains beyond CFG, improving the precision--recall Pareto frontier. Experiments demonstrate the effectiveness of MG across benchmarks. On ImageNet-256, MG improves FID by 36.54\% without CFG and 25.42\% with CFG on average across sampling settings, attaining an FID of 1.553 at 16 sampling steps. Evaluations on large flow-based models, including Stable Diffusion 3 and FLUX.1-dev, further confirm improvements across standard metrics.
\end{abstract}

\section{Introduction}

Continuous-time generative modeling, including diffusion models~\citep{song2019generative, song2020score, ho2020denoising} and flow-based models~\citep{liu2022flow, liu2022rectified, lipman2022flow, albergo2023stochastic, ma2024sit}, offers a simple and effective framework for high-fidelity image, audio, and video synthesis~\citep{esser2024scaling, flux2024, polyak2024movie, wan2025, kong2024hunyuanvideo, chen-etal-2024-f5tts, mehta2024matcha, wu2025qwenimagetechnicalreport, cao2025hunyuanimage}.
Yet pretrained flow models are rarely used in their raw form. In image generation, samples without guidance often appear diffuse, with blurry textures and limited fine-grained detail, suggesting that these models learn an \textbf{oversmoothed} approximation of the data distribution. This behavior is not unique to flow models and is consistent with a familiar regression-to-the-mean effect in neural prediction: when many plausible outputs exist, learned predictors can average over them, suppressing fine details. In image restoration, such averaging is known to produce oversmoothed results with weak high-frequency texture~\citep{sajjadi2017enhancenet, whang2022deblurring}. Analogous distribution-shaping issues arise in language generation, where decoding methods such as temperature scaling and nucleus sampling reshape or truncate the predictive distribution to avoid degenerate or low-diversity text~\citep{holtzman2019curious}.

\begin{algorithm}[t]
\caption{\textbf{Momentum Guidance}}
\label{alg:momentum-guidance}
\begin{algorithmic}[1]
\Require Trained flow model $\vv_{\theta}(\cdot,t)$;
time grid $\{t_i\}$;
EMA $\beta\in[0,1)$;
weight $\alpha \geq 0$
\State Sample $\mZ_{t_0}\sim\mathcal{N}(0,\mI)$
\State Initialize velocity momentum $\vm_{t_0} \gets \vv_\theta(\mZ_{t_0},t_0)$
\For{$i=0$ to $N-1$}
  \State $\Delta t\gets t_{i+1}-t_i$
  \vspace{0.3ex}

  \State $\vv_{t_i}\gets \vv_{\theta}(\mZ_{t_i},t_i)$
  \vspace{0.3ex}

  \State
  $
    \mZ_{t_{i+1}} \gets \mZ_{t_i}
    + \Delta t \Big[\,\vv_{t_i} +
    \colorbox{orange!12}{\text{$\alpha(\vv_{t_i}-\,\vm_{t_{i}})$}}
    \Big]
  $
  \State  \colorbox{orange!12}{$\vm_{t_{i+1}}\gets (1-\beta)\,\vv_{t_i}+\beta\,\vm_{t_i}$}  \Comment{EMA}
  \vspace{0.3ex}
\EndFor
\State \Return $\mZ_{t_N}$
\end{algorithmic}
\end{algorithm}

In flow and diffusion models, this oversmoothing has two main sources. First, the network predicts conditional statistics such as velocity, clean data, or noise; under mean-squared training objectives, these targets are conditional means and average over multiple plausible outcomes, yielding smoothed estimates of the transport dynamics~\citep{scarvelis2023closed, gao2024flow, kamb2024analytic, biroli2024dynamical}. Second, the exponential moving average (EMA) of model parameters, widely used to reduce visual artifacts and improve sample quality, averages model states along the optimization trajectory and can further smooth the learned velocity field~\citep{izmailov2018averaging, karras2024analyzing, nichol2021improved}. Together, these factors bias pretrained models toward diffuse, low-detail outputs.

\begin{figure*}[t]
  \centering
  \includegraphics[width=\linewidth]
  {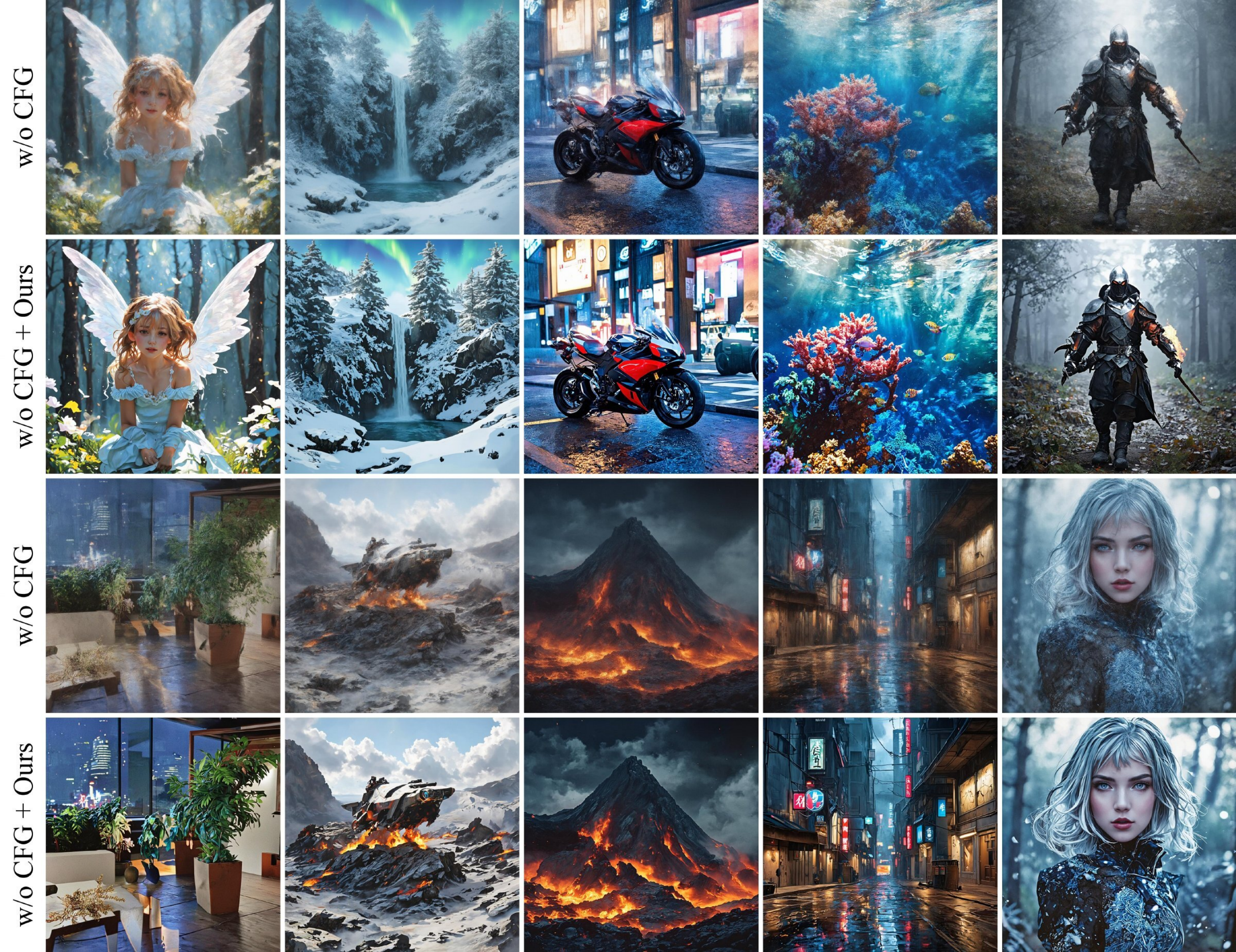}
  \caption{Visual comparison of Momentum Guidance with text-conditioned sampling \textbf{without} CFG on SD3~\citep{esser2024scaling}. Unlike CFG, which requires an additional forward pass through an unconditional branch at every sampling step, MG introduces no extra model evaluations. The generated images show improved quality and finer details (e.g., angel's wings, intricate coral structures), fewer artifacts (e.g., reduced blur in motorcycle reflections), richer visual textures and color variation (e.g., waterfall and volcanic scenes), and more stable object geometry (e.g., clearer facial contours and cleaner edges).}
  \label{fig:cfg1p0cmp}
\end{figure*}

Inference-time guidance mitigates this issue by pushing predictions away from smoother reference estimates. Classifier-free guidance (CFG)~\citep{ho2022classifier, rombach2022high} extrapolates the current conditional prediction away from a smoother unconditional model, while Autoguidance~\citep{karras2024guiding} replaces this unconditional branch with a weaker prediction model, such as an earlier checkpoint or a lower-capacity network, whose outputs tend to be smoother. In this sense, both methods can be viewed as effectively \textbf{de-smoothing} the model's predictions. However, Autoguidance depends on auxiliary checkpoints, which are rarely released for large open models~\citep{flux2024, wu2025qwenimagetechnicalreport}, making it impractical in many settings.

In this work, we introduce \textbf{Momentum Guidance (MG)}, an inference-time technique that uses the ODE trajectory itself to form a smoother velocity reference. MG maintains a velocity momentum, defined as an exponential moving average of past model velocity estimates from earlier, higher-noise states where predictions are intrinsically smoother. Extrapolating the current velocity away from this EMA reference produces the sharpening effect associated with guidance while preserving the standard one-evaluation-per-step cost. MG requires no auxiliary models, no additional network evaluations, and works effectively both with and without CFG\@.

We validate MG across diverse benchmarks. On ImageNet-256~\citep{deng2009imagenet}, MG improved FID by \(36.54\%\) without CFG, and by \(25.42\%\) on top of CFG, achieving an FID of \(1.553\) at 16 sampling steps. Moreover, MG consistently improves the precision--recall Pareto frontier over CFG, achieving trade-offs unattainable by tuning the CFG scale. Evaluations on large flow-based text-to-image models, including Stable Diffusion 3 (SD3)~\citep{esser2024scaling} and FLUX.1-dev~\citep{flux2024}, reveal consistent gains across standard metrics. Due to its simplicity, efficiency, and broad compatibility, MG provides a practical approach to enhance generative quality.

\section{Background}

\subsection{Rectified Flow}
We introduce flow-based generative modeling under the Rectified Flow (RF) framework~\citep{liu2022flow, liu2022rectified}.
Let \(\pi_0\) be a source distribution, typically a Gaussian, and let \(\pi_1=\pi_{\text{data}}\) be the target data distribution. RF defines a linear interpolation
\begin{equation}\label{eq:interpX}
    \mX_t = t \mX_1 + (1 - t)\mX_0, \quad \mX_0\sim\pi_0,\quad \mX_1\sim\pi_1,\quad t \in [0, 1],
\end{equation}
and we denote the marginal distribution of \(\mX_t\) by \(\pi_t\). The corresponding RF velocity field is
\begin{equation}
    \vv^*_t(\vx) = \E_{\mX_0, \mX_1} \left[\mX_1-\mX_0 \mid \mX_t = \vx\right],
\end{equation}
which defines the flow ODE
\begin{equation}\label{eq:odeZ}
    \frac{\mathrm d}{\mathrm dt}\mZ_t = \vv^*_t(\mZ_t), \quad \mZ_0 \sim \pi_0.
\end{equation}

A key property of RF is marginal preservation: if \(\mZ_0\sim\pi_0\), then the ODE solution satisfies \(\mZ_t\sim\pi_t\) for all \(t\in[0,1]\), and thus \(\mZ_1\sim\pi_1\). Therefore, integrating the flow ODE from the source distribution yields samples from the target distribution at \(t=1\). In practice, the exact velocity is approximated by a neural network \(\vv_\theta(\vx,t)\) trained with the mean squared loss
\begin{equation}
    \Ls(\theta) = \E_{\mX_0, \mX_1, t} \left[ \norm{\mX_1-\mX_0 - \vv_\theta(\mX_t, t)}^2 \right],
\end{equation}
where \(t\) is sampled from \([0,1]\). Generation then proceeds by numerically integrating the learned ODE~\citep{song2020denoising, karras2022elucidating}, commonly with the Euler update
\begin{equation}
    \mZ_{t_{i+1}} = \mZ_{t_i} + (t_{i+1} - t_i)\,\vv_\theta(\mZ_{t_i}, t_i).
\end{equation}

\paragraph{Different levels of smoothness in flow marginals.}
With a Gaussian source $\pi_0$, the marginal \(\pi_t\) corresponds to the data distribution smoothed by Gaussian kernels and can be expressed as
\begin{equation}\label{eq:xtdist}
    \pi_t(\vx_t) = \sum_{\vx_1 \in \mathcal{D}_{\text{data}}} \pi_1(\vx_1)\,
    \mathcal{N}\!\left(\vx_t;\,t\vx_1,\,(1 - t)^2 \boldsymbol I\right).
\end{equation}

Smaller values of \(t\) therefore correspond to more strongly smoothed marginals, and during inference \(\mZ_t \sim \pi_t\) evolves toward distributions of decreasing smoothness over time. The velocity field \(\vv_t^*(\vx)\) is linked to the marginal \(\pi_t\) through the score function. In particular,
\begin{equation}
    \nabla_\vx \log \pi_t(\vx) = \frac{t\,\vv^*_t(\vx) - \vx}{1 - t}.
\end{equation}

See, e.g., \citep{lq2024rectifiedflow, hu2025amo, let2025Liu, hu2025improving}. Thus, velocity estimates along the trajectory inherit the same smoothness ordering as the marginals: earlier velocities are associated with smoother distributions, while later velocities correspond to sharper, more data-like distributions.

\subsection{Guidance Methods}

Guidance methods can be viewed as extrapolating a primary velocity away from a smoother reference velocity. This perspective is especially natural in flow models, where smoother velocity fields correspond to smoother marginals and extrapolation acts as an inference-time de-smoothing operation.

\paragraph{Classifier-Free Guidance (CFG).}
Classifier-free guidance (CFG)~\citep{ho2022classifier, rombach2022high} uses the conditional velocity as the primary prediction and the unconditional velocity as the reference:
\begin{equation}
\vv^{\text{CFG}}(\vx, t \mid c) = w\,\vv(\vx, t \mid c) + (1 - w)\,\vv(\vx, t \mid \emptyset),
\end{equation}
where \(w>1\) controls the extrapolation strength. The unconditional branch is smoother because it averages over conditioning variables:
\begin{equation}
\vv(\vx, t \mid \emptyset) = \E_c \!\left[ \vv(\vx, t \mid c) \right],
\end{equation}
where \(c\) may represent class labels, text embeddings, or other attributes. Moving away from this smoother branch improves fidelity and condition alignment, but often reduces diversity~\citep{sadat2023cads, kynkaanniemi2024applying, papalampidi2025dynamic}.

\paragraph{Autoguidance.}
Autoguidance~\citep{karras2024guiding} replaces the unconditional branch with a weaker reference model, typically an earlier checkpoint or a lower-capacity variant:
\begin{equation}
\vv^{\text{Auto}}(\vx, t) = w\,\vv(\vx, t) + (1 - w)\,\vv'(\vx, t),
\end{equation}
where \(\vv'\) tends to produce smoother predictions than the main model. This avoids an unconditional branch, but requires an auxiliary checkpoint, which is often unavailable for large open models~\citep{flux2024, esser2024scaling}, and also increases memory usage.

\section{Momentum Guidance}

\begin{figure*}[t]
  \centering
  \includegraphics[width=\linewidth]{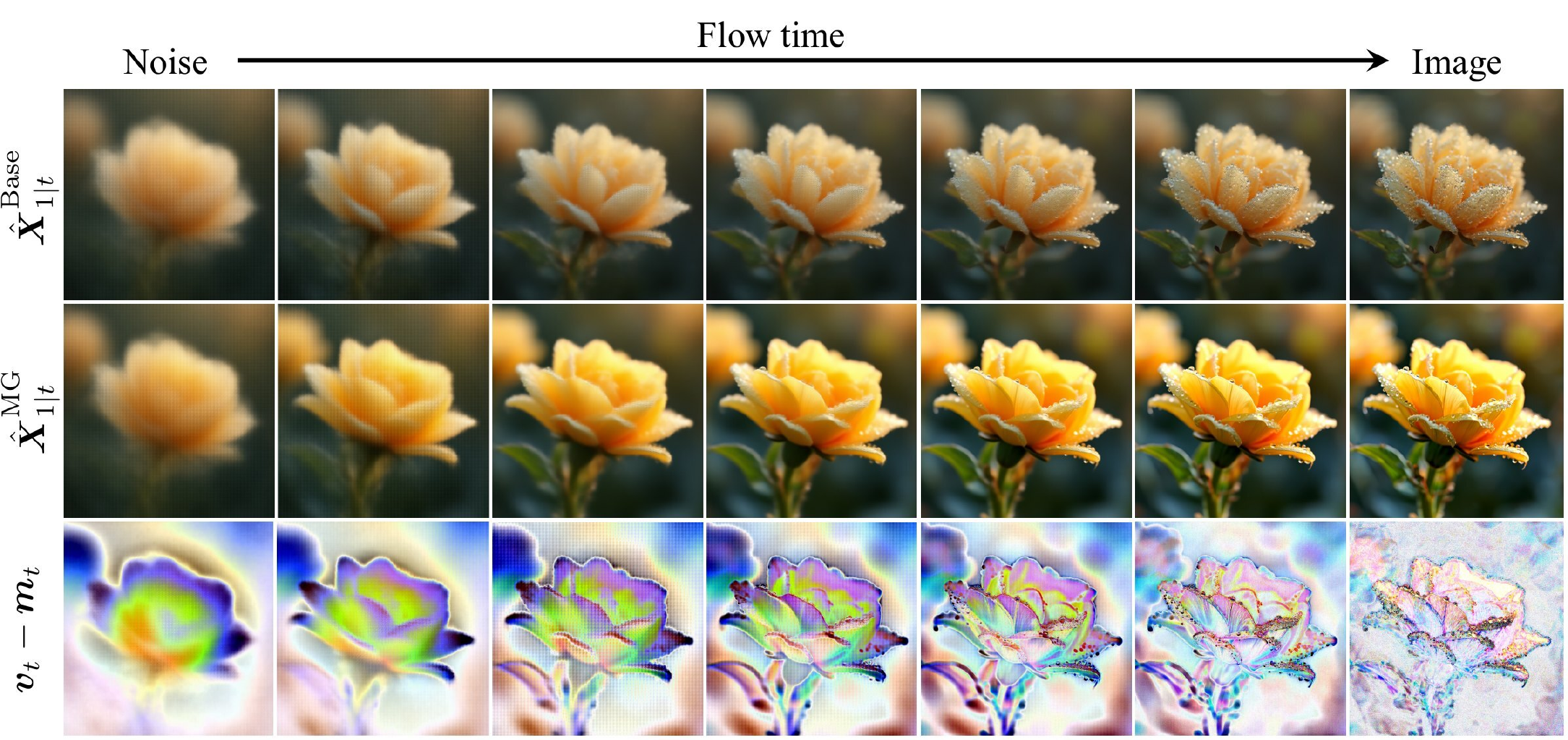}
  \caption{Momentum guidance along a sampling trajectory.
From left to right, flow time increases and the data estimates evolve from blurry previews to a clean image. The first two rows compare the baseline estimates \(\hat{\boldsymbol X}_{1\mid t}^{\text{Base}}\) with the momentum-guided estimates \(\hat{\boldsymbol X}_{1\mid t}^{\text{MG}}\), where MG yields sharper structure, stronger contrast, and clearer fine details.
The third row shows the extrapolation direction \((\boldsymbol v_t - \boldsymbol m_t)\) induced by the velocity momentum, highlighting coarse contours early in the trajectory and fine details near the end.}

  \label{fig:flower}
\end{figure*}
Momentum Guidance (MG) uses the sampler's own trajectory as the reference branch. By RF marginal preservation, earlier points along the trajectory lie on higher-noise, smoother marginals, while later points approach sharper data-like marginals (Eq.~\ref{eq:xtdist}). Thus, past velocities already provide the smoother reference that CFG obtains from an unconditional branch and Autoguidance obtains from an auxiliary model. MG stores this reference as a \emph{velocity momentum}: an exponential moving average of past model velocity estimates, analogous to momentum mechanisms in optimization~\citep{chen2024lion,sutskever2013importance,kingma2014adam,nguyen2024memory,liang2024memory,liang2024cautious,nguyen2025improving,chen2025muon,liu2024communication,peng2026demodecoupledmomentumoptimization,chen2025cautious}. The sampler then extrapolates the current velocity away from this EMA reference, yielding plug-and-play guidance without extra network evaluations.

Let \(\mZ_t\) denote the RF ODE state, and let \(\vm_t\) denote the velocity momentum, i.e., the EMA reference velocity accumulated along the sampling trajectory. We initialize $\mZ_0 \sim \mathcal{N}(0,\mI)$ and set $\vm_{t_0} = \vv_{\theta}(\mZ_{t_0}, t_0)$. At each timestep \(t_i\), given the current model velocity \(\vv_{t_i} \coloneqq \vv_\theta(\mZ_{t_i}, t_i)\), the velocity momentum is updated as
\begin{equation}
    \vm_{t_{i+1}} = (1 - \beta)\,\vv_{t_i} + \beta\,\vm_{t_i},
\end{equation}
where $\beta$ controls the decay of the velocity history. We then update the sample using an extrapolated velocity:
\begin{equation}
    \mZ_{t_{i+1}} = \mZ_{t_i}
    + \Delta t \Big[\,\vv_{t_i} +
    \alpha\big(\vv_{t_i} - \vm_{t_{i}}\big)\Big],
    \label{eq:mg_update}
\end{equation}
with $\Delta t = t_{i+1}-t_i$ and $\alpha > 0$ governing the extrapolation strength toward sharper distributions. The final sample is obtained at time $t_N = 1$.

\paragraph{Memory and computation overhead.}
Momentum Guidance does not change the number of function evaluations. Each step reuses the model velocity already computed by the base sampler, then applies the extrapolation update in Eq.~\ref{eq:mg_update}. The only additional state is the velocity momentum $\vm_{t_i}$, an EMA buffer with the same shape as the flow state $\mZ_{t_i}$. This overhead is negligible compared with model parameters and intermediate activations. For example, ImageNet-256 models~\citep{peebles2023scalable} with an SD encoder~\citep{rombach2022high} use latents of size \(32\times 32\times 4\), while high-resolution models such as FLUX.1-dev~\citep{flux2024} use latents on the order of \(128\times 128 \times 16\) for \(1024^2\) images.

\paragraph{Understanding Momentum Guidance.}

To make the effect of MG visible along the ODE, we examine the implied clean-data prediction at each inference step,
\begin{equation}
\hat{\boldsymbol{X}}_{1\mid t}
\!=\!\mathbb{E}\!\left[\boldsymbol X_1\! \mid\! \boldsymbol X_t=\boldsymbol x_t\right]
\!=\! \boldsymbol{x}_t + (1-t)\,\boldsymbol{v}_\theta(\boldsymbol{x}_t,t),
\end{equation}
i.e., the conditional data mean implied by the learned velocity field. It provides a direct image-space view of the evolving sample. This diagnostic is not tied to a particular network parameterization. In data-prediction models it is produced directly, while in velocity-prediction models it follows from the RF identity above; the two parameterizations are equivalent up to a time-dependent training weight~\citep{let2025Liu, gao2025diffusionmeetsflow}. Visualizing \(\hat{\boldsymbol{X}}_{1\mid t}\) provides a common view of how guidance reshapes the trajectory toward the data distribution.

Figure~\ref{fig:flower} compares the FLUX.1-dev Euler sampler with \(\text{CFG}\; \omega=1.5\) against MG with \(\alpha=0.6\) and \(\beta=0.8\). Relative to the baseline estimates \(\hat{\boldsymbol X}_{1\mid t}^{\text{Base}}\), the MG estimates \(\hat{\boldsymbol X}_{1\mid t}^{\text{MG}}\) develop clearer object structure and more stable color earlier in the trajectory. The bottom row shows the extrapolation direction \((\boldsymbol v_t-\boldsymbol m_t)\). Since \(\boldsymbol m_t\) aggregates earlier, smoother velocities, this difference isolates information newly emerging beyond the EMA reference: it first aligns with coarse object geometry, then concentrates on high-frequency details such as petal boundaries and dew droplets as the flow approaches the data distribution.

\section{Experiments}

We evaluate Momentum Guidance across three settings. On ImageNet~\citep{deng2009imagenet}, we conduct a systematic ablation over the guidance weight $\alpha$, EMA decay $\beta$, and sampling budget. MG consistently reduces FID and improves sample quality across these settings. We then apply MG to large-scale text-to-image models, including FLUX.1-dev~\citep{flux2024} and Stable Diffusion 3~\citep{esser2024scaling}, and further extend it to text-to-video generation with HunyuanVideo~\citep{kong2024hunyuanvideo} evaluated by VBench~\citep{huang2023vbench}. Across image and video domains, MG improves visual fidelity and structural coherence, with extended qualitative comparisons provided in the Appendix.

\begin{table}[!ht]
    \centering
    \caption{Comparison of CFG and our Momentum Guidance across different CFG scales \(w\) at different NFEs. When \(w=1\), our method corresponds to MG without CFG, while the other settings represent MG applied on top of CFG\@.}
    \label{tab:cfg_vs_ours}
    \begingroup
    \renewcommand{\arraystretch}{1.12}
    \footnotesize
    \begin{adjustbox}{max width=\textwidth,center}
    \begin{tabular}{cccccc}
        \toprule
        \multirow{2}{*}{$w$} & \multirow{2}{*}{Method} &
        \multicolumn{4}{c}{NFE = 16 / 32 / 64} \\
        \cmidrule(lr){3-6}
        & & FID-50K $\downarrow$ & IS $\uparrow$ & Precision $\uparrow$ & Recall $\uparrow$ \\
        \midrule
        \multirow{2}{*}{1.0}
            & w/o CFG  & 7.76 / 5.57 / 4.75 & 140.89 / 156.10 / 163.50 & 0.70 / 0.72 / 0.72 & 0.65 / \textbf{0.67} / \textbf{0.67} \\
            &\cellcolor{lightpink} Ours &\cellcolor{lightpink} \textbf{4.46} / \textbf{3.58} / \textbf{3.26} &\cellcolor{lightpink} \textbf{165.85} / \textbf{176.10} / \textbf{179.66} &\cellcolor{lightpink} \textbf{0.73} / \textbf{0.74} / \textbf{0.74} &\cellcolor{lightpink} \textbf{0.66} / \textbf{0.67} / \textbf{0.67} \\
        \midrule
        \multirow{2}{*}{1.2}
            & CFG  & 3.26 / 2.20 / 1.89 & 212.83 / 230.71 / 239.56 & 0.78 / 0.79 / 0.79 & 0.60 / \textbf{0.62} / \textbf{0.62} \\
                &\cellcolor{lightpink} Ours &\cellcolor{lightpink} \textbf{2.00} / \textbf{1.71} / \textbf{1.60} &\cellcolor{lightpink} \textbf{238.29} / \textbf{250.89} / \textbf{254.60} &\cellcolor{lightpink} \textbf{0.80} / \textbf{0.81} / \textbf{0.81} &\cellcolor{lightpink} \textbf{0.61} / \textbf{0.62} / \textbf{0.62} \\
        \midrule
        \multirow{2}{*}{1.4}
            & CFG  & 2.38 / 2.04 / 2.03 & 275.06 / 293.03 / 301.29 & \textbf{0.83} / \textbf{0.84} / \textbf{0.84} & 0.57 / 0.58 / 0.58 \\
            &\cellcolor{lightpink} Ours &\cellcolor{lightpink} \textbf{1.85} / \textbf{1.90} / \textbf{1.99} &\cellcolor{lightpink} \textbf{288.65} / \textbf{300.71} / \textbf{306.02} &\cellcolor{lightpink} 0.82 / \textbf{0.84} / \textbf{0.84} &\cellcolor{lightpink} \textbf{0.60} / \textbf{0.59} / \textbf{0.59} \\
        \midrule
        \multirow{2}{*}{1.6}
            & CFG  & 3.13 / 3.17 / 3.34 & 325.55 / 340.88 / 348.87 & \textbf{0.86} / \textbf{0.87} / \textbf{0.87} & 0.52 / 0.54 / 0.54 \\
            &\cellcolor{lightpink} Ours &\cellcolor{lightpink} \textbf{2.62} / \textbf{2.89} / \textbf{3.17} &\cellcolor{lightpink} \textbf{330.08} / \textbf{342.27} / \textbf{349.04} &\cellcolor{lightpink} 0.85 / 0.85 / 0.85 &\cellcolor{lightpink} \textbf{0.56} / \textbf{0.56} / \textbf{0.56} \\
        \midrule
        \multirow{2}{*}{1.8}
            & CFG  & 4.48 / 4.76 / 4.99 & \textbf{363.56} / \textbf{377.39} / \textbf{383.60} & \textbf{0.89} / \textbf{0.89} / \textbf{0.89} & 0.49 / 0.49 / 0.49 \\
            & \cellcolor{lightpink} Ours & \cellcolor{lightpink} \textbf{3.49} / \textbf{3.96} / \textbf{4.62} & \cellcolor{lightpink} 353.77 / 370.72 / 382.04 & \cellcolor{lightpink} 0.85 / 0.86 / 0.87 & \cellcolor{lightpink} \textbf{0.54} / \textbf{0.53} / \textbf{0.51} \\
        \midrule
        \multirow{2}{*}{2.0}
            & CFG  & 5.94 / 6.36 / 6.62 & \textbf{392.50} / \textbf{403.22} / \textbf{408.68} & \textbf{0.90} / \textbf{0.90} / \textbf{0.90} & 0.45 / 0.46 / 0.46 \\
            & \cellcolor{lightpink} Ours & \cellcolor{lightpink} \textbf{4.62} / \textbf{5.27} / \textbf{6.08} & \cellcolor{lightpink} 382.79 / 397.44 / 407.30 & \cellcolor{lightpink} 0.87 / 0.88 / 0.89 & \cellcolor{lightpink} \textbf{0.51} / \textbf{0.49} / \textbf{0.48} \\
        \bottomrule
    \end{tabular}
    \end{adjustbox}
    \endgroup
\end{table}

\begin{table}[t]
  \centering
  \caption{Effect of applying a CFG interval schedule~\citep{kynkaanniemi2024applying} on ImageNet-256.
\textit{CFG-int} activates CFG only on $t\in[0.125,1]$, while Momentum Guidance is applied over the full interval $t\in[0,1]$.
We report FID/IS/Precision/Recall on 50K samples.
$\dagger$ denotes methods that use additional vision foundation models beyond the base flow model~\citep{zheng2025diffusion}.
We implement the baselines ADG~\citep{jin2025angle} and CFG++~\citep{chung2025cfg}, search over the suggested hyperparameters, and report the best FID\@.}
  \label{tab:ema_euler_cfg_interval}
  \begingroup
  \setlength{\tabcolsep}{5pt}
  \renewcommand{\arraystretch}{1.08}
  \footnotesize

  \newcommand{\HL}[1]{\cellcolor{pink!12}#1}
  \newcolumntype{Y}{>{\centering\arraybackslash}X}

  \begin{tabularx}{\textwidth}{c Y c c c c}
    \toprule
    NFE & Method & FID-50K $\downarrow$ & IS $\uparrow$ & Precision $\uparrow$ & Recall $\uparrow$ \\
    \midrule

    \multirow{8}{*}{16}
      & CFG-int ($\omega{=}1.4$)                                   & 2.352 & 249.85 & 0.791 & 0.612 \\
      & \HL{CFG-int + MG ($\omega{=}1.4$)}                         & \HL{\textbf{1.553}} & \HL{268.03} & \HL{0.799} & \HL{0.636} \\
    \addlinespace[0.25em]
      & CFG-int ($\omega{=}1.6$)                                   & 1.993 & 291.80 & 0.819 & 0.594 \\
      & \HL{CFG-int + MG ($\omega{=}1.6$)}                         & \HL{\underline{1.638}} & \HL{\textbf{306.57}} & \HL{0.822} & \HL{0.611} \\
    \addlinespace[0.25em]
      & ADG ($\omega{=}1.2$)                                       & 3.150 & 214.76 & 0.777 & 0.610 \\
      & ADG ($\omega{=}1.4$)                                       & 2.324 & 275.28   & 0.827 & 0.570 \\
    \addlinespace[0.25em]
      & CFG++ ($\omega{=}0.4$)                                     & 3.223 & 226.98    & 0.736 & 0.668 \\
      & CFG++ ($\omega{=}0.6$)                                     & 2.620 & 368.80    & 0.842 & 0.572 \\
    \cmidrule(lr){1-6}

    \multirow{8}{*}{32}
      & CFG-int ($\omega{=}1.4$)                                   & 1.642 & 264.75 & 0.800 & 0.626 \\
      & \HL{CFG-int + MG ($\omega{=}1.4$)}                         & \HL{\textbf{1.408}} & \HL{274.99} & \HL{0.801} & \HL{0.633} \\
    \addlinespace[0.25em]
      & CFG-int ($\omega{=}1.6$)                                   & 1.639 & 305.82 & 0.822 & 0.605 \\
      & \HL{CFG-int + MG ($\omega{=}1.6$)}                         & \HL{\underline{1.581}} & \HL{\textbf{308.74}} & \HL{0.822} & \HL{0.613} \\
    \addlinespace[0.25em]
      & ADG ($\omega{=}1.2$)                                       & 2.160 & 231.57    & 0.790 & 0.628 \\
      & ADG ($\omega{=}1.4$)                                       & 2.000 & 294.47    & 0.835 & 0.582 \\
    \addlinespace[0.25em]
      & CFG++ ($\omega{=}0.3$)                                     & 1.714 & 342.90    & 0.809 & 0.624 \\
      & CFG++ ($\omega{=}0.4$)                                     & 4.104 & 419.78    & 0.857 & 0.547 \\
    \cmidrule(lr){1-6}

    \multirow{8}{*}{64}
      & CFG-int ($\omega{=}1.4$)                                   & 1.462 & 271.34 & 0.803 & 0.625 \\
      & \HL{CFG-int + MG ($\omega{=}1.4$)}                         & \HL{\textbf{1.380}} & \HL{277.73} & \HL{0.802} & \HL{0.630} \\
    \addlinespace[0.25em]
      & CFG-int ($\omega{=}1.6$)                                   & 1.612 & 311.17 & 0.824 & 0.608 \\
      & \HL{CFG-int + MG ($\omega{=}1.6$)}                         & \HL{1.618} & \HL{\textbf{314.33}} & \HL{0.824} & \HL{0.613} \\
    \addlinespace[0.25em]
      & ADG ($\omega{=}1.2$)                                       & 1.846 & 240.03    & 0.793 & 0.634 \\
      & ADG ($\omega{=}1.4$)                                       & 1.997 & 302.36    & 0.837 & 0.585 \\
    \addlinespace[0.25em]
      & CFG++ ($\omega{=}0.2$)                                     & 3.390 & 405.26    & 0.838 & 0.573 \\
      & CFG++ ($\omega{=}0.3$)                                     & 7.123 & 468.43    & 0.875 & 0.470 \\
    \midrule

    \multicolumn{6}{c}{\textit{Baselines using \textbf{additional} vision foundation models ($\dagger$)}}\\
    \midrule

    \multirow{2}{*}{50 / 2}
      & RAE$^\dagger$ ($\omega{=}1.0$)              & 1.535 & 241.56 & 0.791 & 0.644 \\
          & \HL{RAE$^\dagger$ + MG ($\omega{=}1.0$)}  & \HL{\textbf{1.376}} & \HL{242.83} & \HL{0.790} & \HL{0.642} \\
    \addlinespace[0.25em]
    \multirow{2}{*}{50}
          & RAE$^\dagger$ ($\omega{=}1.5$)              & 3.723 & 359.44 & 0.866 & 0.551 \\
          & \HL{RAE$^\dagger$ + MG ($\omega{=}1.5$)}  & \HL{3.279} & \HL{346.74} & \HL{0.855} & \HL{0.576} \\
    \bottomrule
  \end{tabularx}
  \endgroup
\end{table}

\FloatBarrier

\subsection{Main results}

\paragraph{Results on ImageNet} We evaluate Momentum Guidance on ImageNet at \(256\times256\) using the official Rectified Flow codebase~\citep{let2025Liu} with an improved DiT-XL architecture~\citep{yao2025reconstruction}. Implementation and training details are provided in the Appendix. Unless otherwise specified, we use the standard Euler sampler on a uniformly discretized time grid. For MG, we include the unbiased EMA correction~\citep{kingma2014adam} and momentum normalization. Motivated by Guidance interval~\citep{kynkaanniemi2024applying}, we also sweep the MG application interval over \([0.1,0.6]\), \([0.1,0.7]\), \([0.2,0.6]\), and \([0.0,1.0]\).
When combined with CFG, MG treats the CFG-adjusted velocity as the base velocity estimate and keeps CFG active at every timestep. We report standard metrics: Fr\'echet Inception Distance (FID)~\citep{heusel2017gans}, Inception Score (IS)~\citep{salimans2016improved}, and Precision/Recall (P/R)~\citep{sajjadi2018assessing}. For each CFG scale and NFE budget, we select \((\alpha,\beta)\) by grid search on FID-10K and report the best configurations on FID-50K.

Table~\ref{tab:cfg_vs_ours} shows that Momentum Guidance consistently reduces FID across sampling budgets and guidance strengths. The strongest result is achieved at CFG $=1.2$ with $64$ NFEs, where MG obtains the best FID among all configurations in Table~\ref{tab:cfg_vs_ours}. Unlike simply increasing CFG, which improves precision at the cost of recall, MG improves sample quality without degrading recall. Even without CFG, MG reduces FID by $36.54\%$ on average while requiring only one network evaluation per step, and it yields a $25.42\%$ average reduction at CFG $=1.2$. We further compare against guidance-related baselines in Table~\ref{tab:ema_euler_cfg_interval}, including a CFG interval schedule that activates CFG only on $t\in[0.125,1]$~\citep{kynkaanniemi2024applying}. MG remains competitive across settings and reaches an FID of 1.553 with only 16 NFEs.

\paragraph{Results on Text-to-Image Generation} We next evaluate Momentum Guidance on large-scale text-to-image generation with FLUX.1-dev and Stable Diffusion 3, using CFG as the baseline. For each model, we generate 3,200 images at $1024\times1024$ resolution from prompts in the HPSv2 benchmark~\citep{wu2023human}, and evaluate the resulting image-prompt pairs with HPSv2.1 and ImageReward~\citep{xu2023imagereward}. Inference uses the default Euler sampler together with the default discretization schedule of each model. We tune only $\alpha$ and $\beta$, apply MG with normalized momentum at all timesteps, and do not restrict it to an interval as in ImageNet. Tables~\ref{tab:sd3_results} and~\ref{tab:flux_results} show that MG consistently improves HPSv2.1 over vanilla CFG on both models, while also improving ImageReward in most configurations with only minor drops at a few CFG values.

\begin{table}[t]
\centering
\caption{Results on SD3 with 28 sampling steps. Across all CFG scales, our method improves HPSv2.1 and ImageReward over the baseline.}
\label{tab:sd3_results}
\setlength{\tabcolsep}{3.2pt}
\renewcommand{\arraystretch}{0.95}
\footnotesize
\begin{adjustbox}{max width=\columnwidth}
\begin{tabular}{lcccccccc}
\toprule
\multirow{2}{*}{Metrics} &
\multirow{2}{*}{Method} &
\multicolumn{7}{c}{CFG} \\
\cmidrule(lr){3-9}
& & 1 & 2 & 3 & 4 & 5 & 6 & 7 \\
\midrule
\multirow{2}{*}{HPSv2.1}
& CFG  & 22.87 & 27.99 & 29.38 & 29.98 & 30.22 & 30.39 & 30.41 \\
& \cellcolor{lightpink} Ours
& \cellcolor{lightpink}\textbf{27.37}
& \cellcolor{lightpink}\textbf{29.78}
& \cellcolor{lightpink}\textbf{30.34}
& \cellcolor{lightpink}\textbf{30.43}
& \cellcolor{lightpink}\textbf{30.62}
& \cellcolor{lightpink}\textbf{30.59}
& \cellcolor{lightpink}\textbf{30.56} \\
\midrule
\multirow{2}{*}{IR}
& CFG  & -0.093 & 0.801 & 0.988 & 1.059 & 1.099 & 1.117 & 1.114 \\
& \cellcolor{lightpink} Ours
& \cellcolor{lightpink}\textbf{0.395}
& \cellcolor{lightpink}\textbf{0.926}
& \cellcolor{lightpink}\textbf{1.046}
& \cellcolor{lightpink}\textbf{1.088}
& \cellcolor{lightpink}\textbf{1.111}
& \cellcolor{lightpink}\textbf{1.118}
& \cellcolor{lightpink}\textbf{1.120} \\
\bottomrule
\end{tabular}
\end{adjustbox}
\end{table}

\begin{table}[t]
\centering
\caption{Results on FLUX.1-dev with 50 sampling steps. Our method improves perceptual quality across CFG scales.}
\label{tab:flux_results}
\setlength{\tabcolsep}{3.2pt}
\renewcommand{\arraystretch}{0.95}
\footnotesize
\begin{adjustbox}{max width=\columnwidth}
\begin{tabular}{lcccccccc}
\toprule
\multirow{2}{*}{Metrics} &
\multirow{2}{*}{Method} &
\multicolumn{7}{c}{CFG} \\
\cmidrule(lr){3-9}
& & 1 & 1.5 & 2 & 2.5 & 3 & 3.5 & 4 \\
\midrule
\multirow{2}{*}{HPSv2.1}
& CFG & 24.40 & 29.33 & 30.75 & 31.09 & 31.28 & 31.40 & 31.45 \\
& \cellcolor{lightpink} Ours
& \cellcolor{lightpink}\textbf{24.80}
& \cellcolor{lightpink}\textbf{29.90}
& \cellcolor{lightpink}\textbf{30.82}
& \cellcolor{lightpink}\textbf{31.13}
& \cellcolor{lightpink}\textbf{31.29}
& \cellcolor{lightpink}\textbf{31.41}
& \cellcolor{lightpink}\textbf{31.47} \\
\midrule
\multirow{2}{*}{IR}
& CFG & 0.345 & 0.912 & 1.048 & 1.075 & 1.094 & \textbf{1.117} & \textbf{1.118} \\
& \cellcolor{lightpink} Ours
& \cellcolor{lightpink}\textbf{0.391}
& \cellcolor{lightpink}\textbf{0.935}
& \cellcolor{lightpink}\textbf{1.054}
& \cellcolor{lightpink}\textbf{1.077}
& \cellcolor{lightpink}\textbf{1.096}
& \cellcolor{lightpink}1.115
& \cellcolor{lightpink}1.116 \\
\bottomrule
\end{tabular}
\end{adjustbox}
\end{table}

\FloatBarrier

\paragraph{Results on Text-to-Video Generation} We further test whether Momentum Guidance transfers to text-to-video generation using HunyuanVideo~\citep{kong2024hunyuanvideo}. Following the standard VBench protocol~\citep{huang2023vbench}, we generate one 540p video with 65 frames for each prompt under a fixed inference budget and use the same random seed 42 for all methods. We apply MG with a fixed setting $\alpha{=}0.6$, $\beta{=}0.2$ without exhaustive grid search. Table~\ref{tab:vbench_overall_540p} summarizes the official VBench aggregate scores, where MG improves the total score from 0.816 to 0.819. Table~\ref{tab:vbench_dims_540p} gives the per-dimension breakdown, showing that MG improves content richness and perceptual quality while remaining comparable on consistency and temporal stability. No prompt-specific tuning or additional network evaluations are used.

\begin{table}[!h]
\centering
\caption{Overall VBench scores (quality/semantic/total) corresponding to Table~\ref{tab:vbench_dims_540p}, computed following the official VBench implementation. Higher is better.}
\label{tab:vbench_overall_540p}

\setlength{\tabcolsep}{6pt}
\renewcommand{\arraystretch}{1.08}
\footnotesize

\begin{tabular}{lccc}
\toprule
\textbf{Method} & \textbf{quality} & \textbf{semantic} & \textbf{total} \\
\midrule
Baseline & 0.842 & 0.716 & 0.816 \\
\rowcolor{lightpink}
Ours     & \textbf{0.845} & \textbf{0.717} & \textbf{0.819} \\
\bottomrule
\end{tabular}
\end{table}

\begin{table*}[t]
\centering
\caption{VBench~\citep{huang2023vbench} per-dimension comparison on HunyuanVideo~\citep{kong2024hunyuanvideo} between Baseline and Momentum Guidance ($\alpha{=}0.6$, $\beta{=}0.2$) at 540p resolution. Higher is better.}
\label{tab:vbench_dims_540p}

\setlength{\tabcolsep}{4.5pt}
\renewcommand{\arraystretch}{1.12}
\small

\begin{adjustbox}{max width=\textwidth,center}
\begin{tabular}{lcccccccc}
\toprule
\textbf{\shortstack{Method\\\phantom{x}}} & \textbf{\shortstack{Subject\\Cons.}} & \textbf{\shortstack{Motion\\Smooth.}} & \textbf{\shortstack{Dynamic\\Degree}} & \textbf{\shortstack{Bg\\Cons.}} & \textbf{\shortstack{Scene\\\phantom{x}}} & \textbf{\shortstack{Overall\\Cons.}} & \textbf{\shortstack{Aesthetic\\Quality}} & \textbf{\shortstack{Imaging\\Quality}} \\
\midrule
Baseline & \textbf{0.962} & \textbf{0.992} & 0.639 & 0.974 & \textbf{0.387} & \textbf{0.266} & 0.623 & 0.653 \\
\rowcolor{lightpink}
Ours     & \textbf{0.962} & 0.991 & \textbf{0.653} & \textbf{0.975} & 0.356 & 0.265 & \textbf{0.624} & \textbf{0.668} \\
\bottomrule
\end{tabular}
\end{adjustbox}

\vspace{6pt}

\begin{adjustbox}{max width=\textwidth,center}
\begin{tabular}{lcccccccc}
\toprule
\textbf{\shortstack{Method\\\phantom{x}}} & \textbf{\shortstack{Multiple\\Objects}} & \textbf{\shortstack{Object\\Class}} & \textbf{\shortstack{Color\\\phantom{x}}} & \textbf{\shortstack{Spatial\\Relation}} & \textbf{\shortstack{Temporal\\Flicker}} & \textbf{\shortstack{Temporal\\Style}} & \textbf{\shortstack{Appearance\\Style}} & \textbf{\shortstack{Human\\Action}} \\
\midrule
Baseline & 0.679 & 0.731 & 0.871 & \textbf{0.695} & \textbf{0.991} & \textbf{0.241} & \textbf{0.192} & \textbf{0.930} \\
\rowcolor{lightpink}
Ours     & \textbf{0.708} & \textbf{0.765} & \textbf{0.885} & 0.681 & \textbf{0.991} & 0.240 & \textbf{0.192} & 0.920 \\
\bottomrule
\end{tabular}
\end{adjustbox}

\end{table*}

\FloatBarrier

\subsection{Ablations}

\begin{figure*}[t]
  \centering
  \begin{subfigure}[t]{0.32\textwidth}
    \centering
    \includegraphics[width=\textwidth]{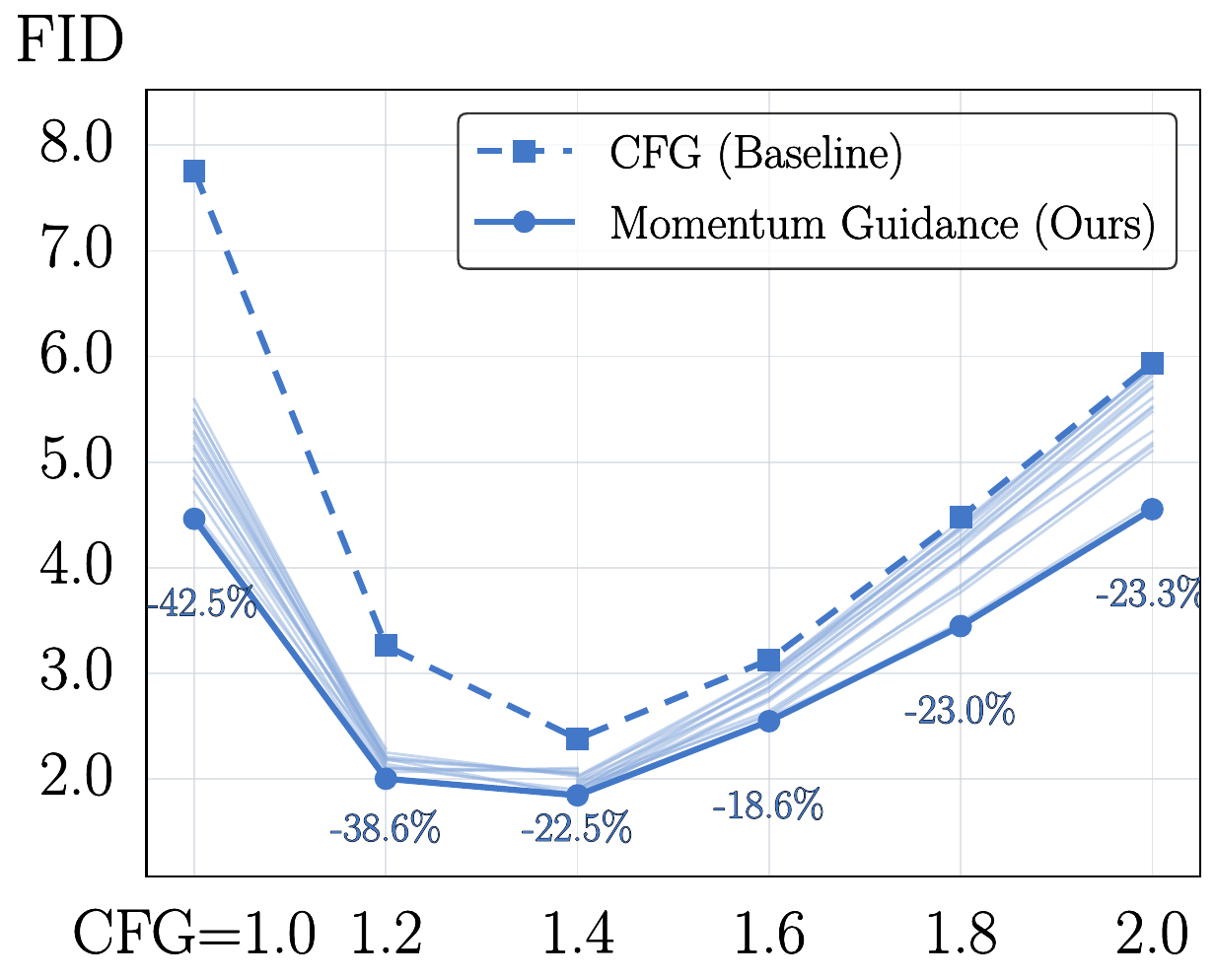}
    \subcaption*{${\textit{NFE}} = 16$}
  \end{subfigure}\hfill
  \begin{subfigure}[t]{0.32\textwidth}
    \centering
    \includegraphics[width=\textwidth]{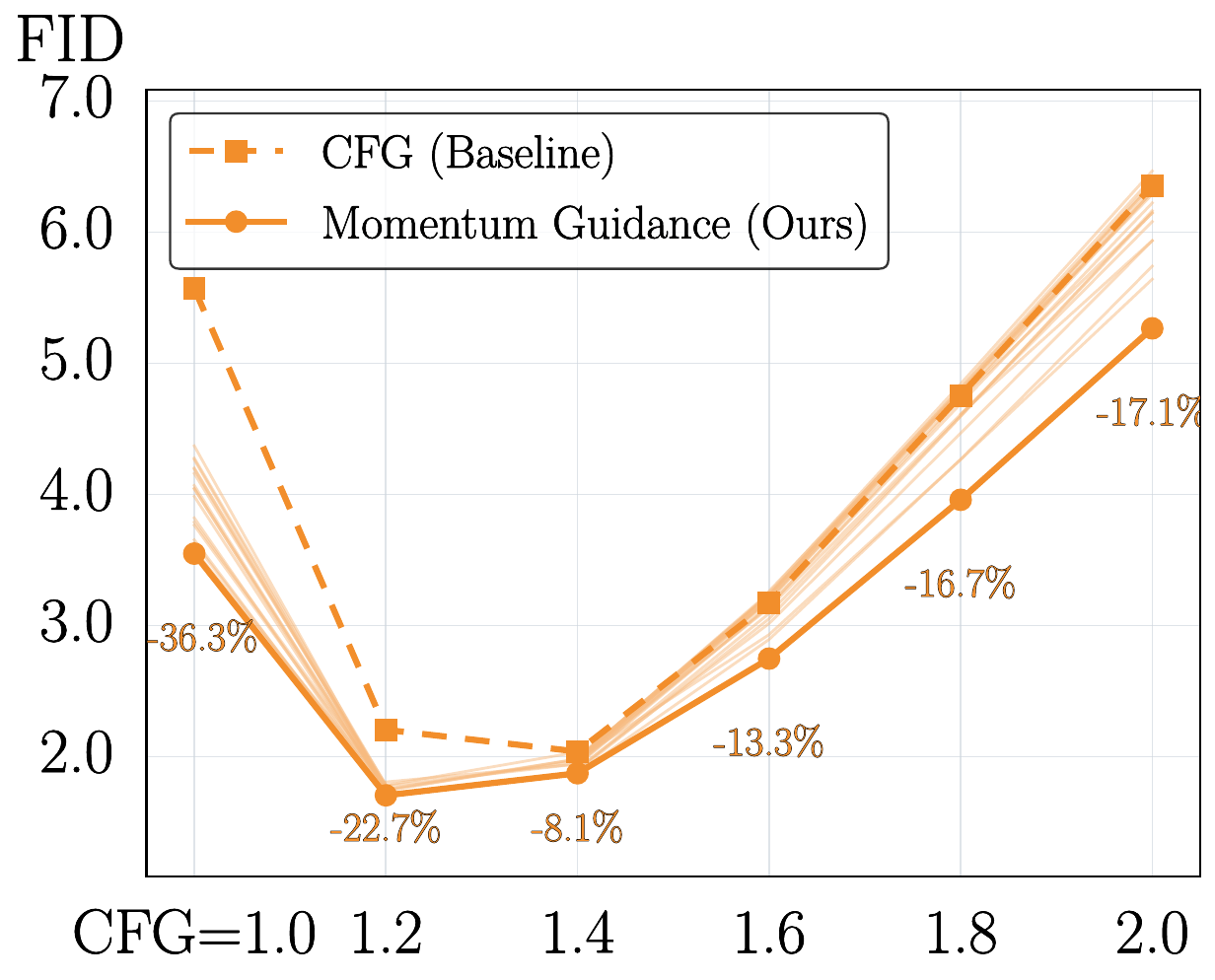}
    \subcaption*{${\textit{NFE}} = 32$}
  \end{subfigure}\hfill
  \begin{subfigure}[t]{0.32\textwidth}
    \centering
    \includegraphics[width=\textwidth]{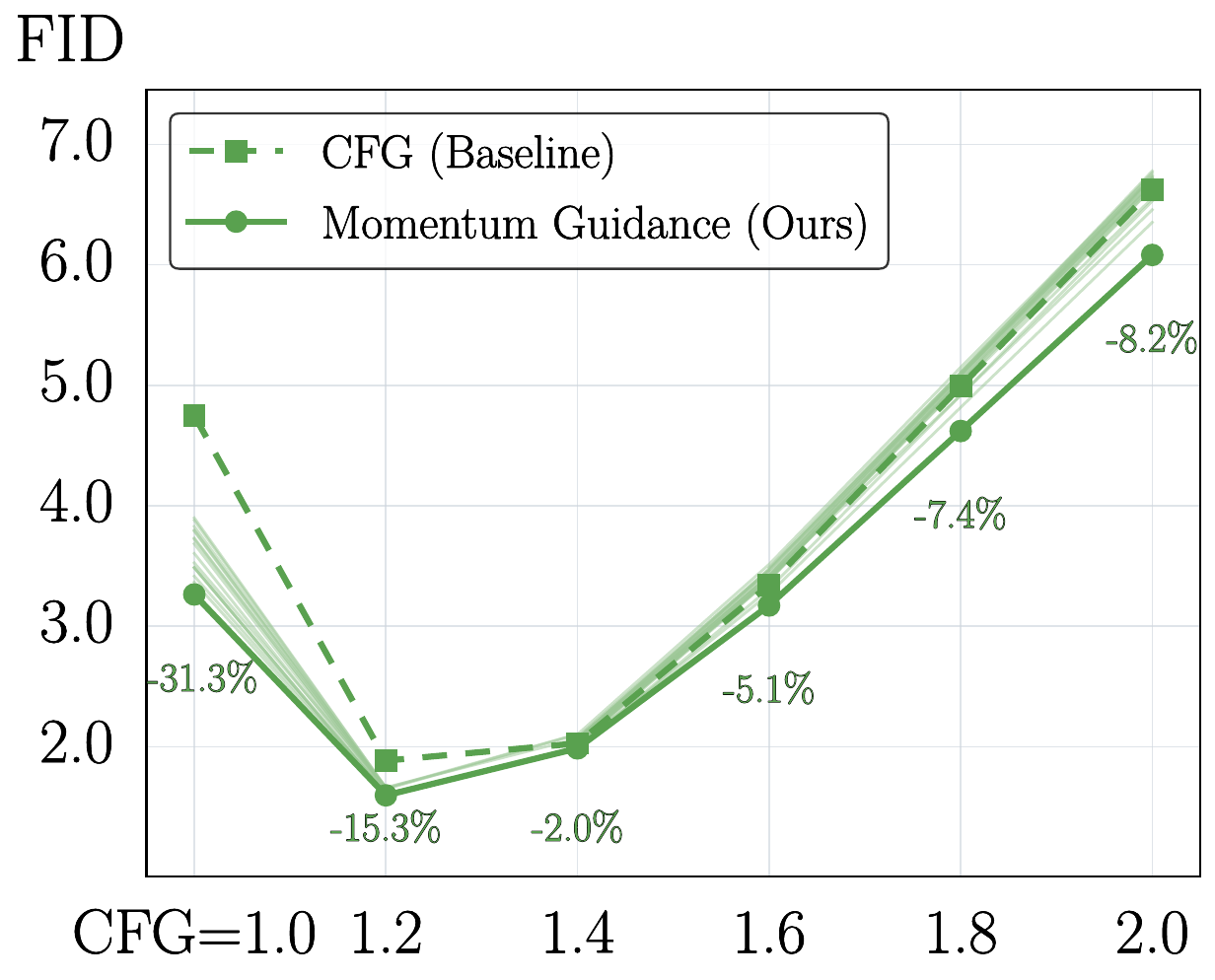}
    \subcaption*{${\textit{NFE}} = 64$}
  \end{subfigure}\\

  \begin{subfigure}[t]{0.32\textwidth}
    \centering
    \includegraphics[width=\textwidth]{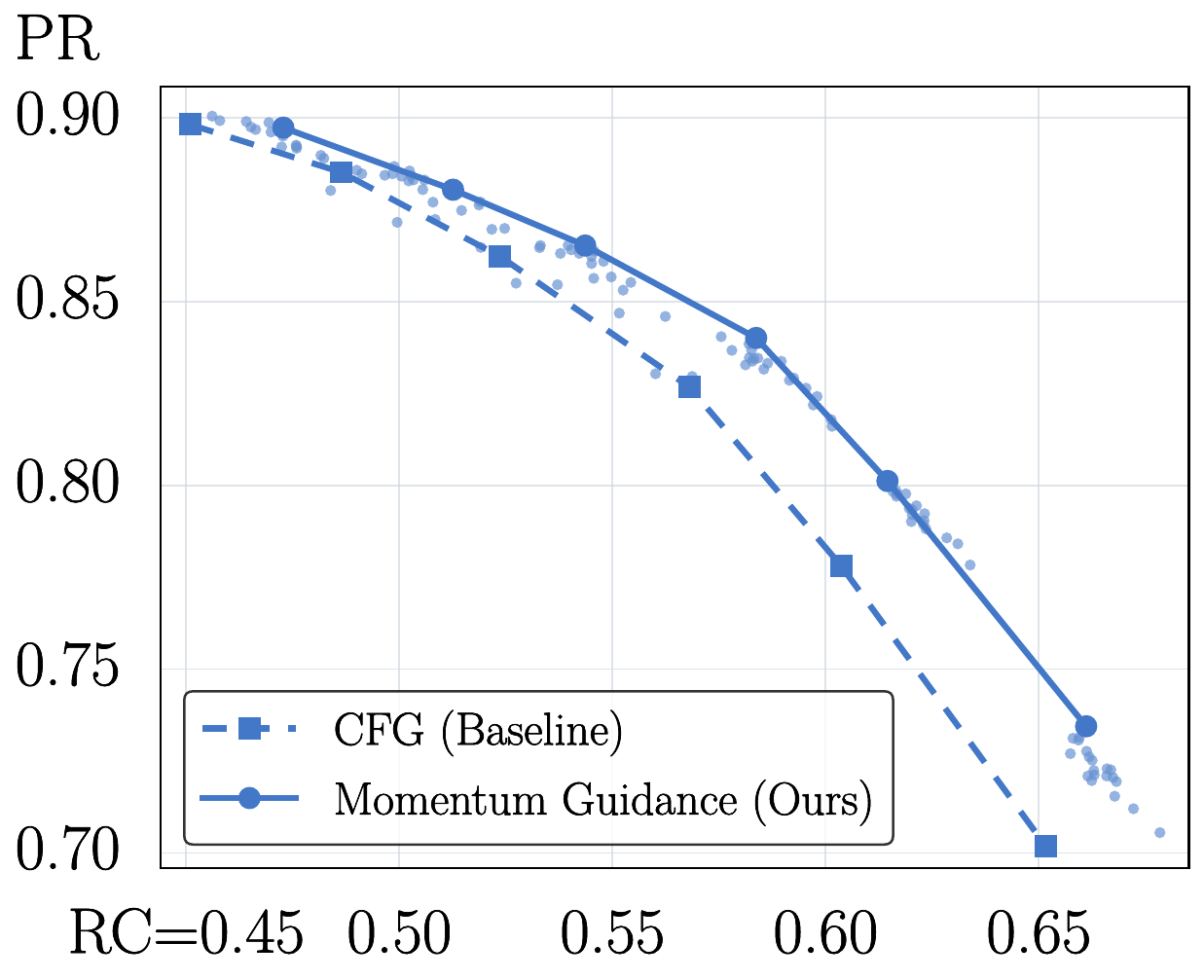}
    \subcaption*{${\textit{NFE}} = 16$}
  \end{subfigure}\hfill
  \begin{subfigure}[t]{0.32\textwidth}
    \centering
    \includegraphics[width=\textwidth]{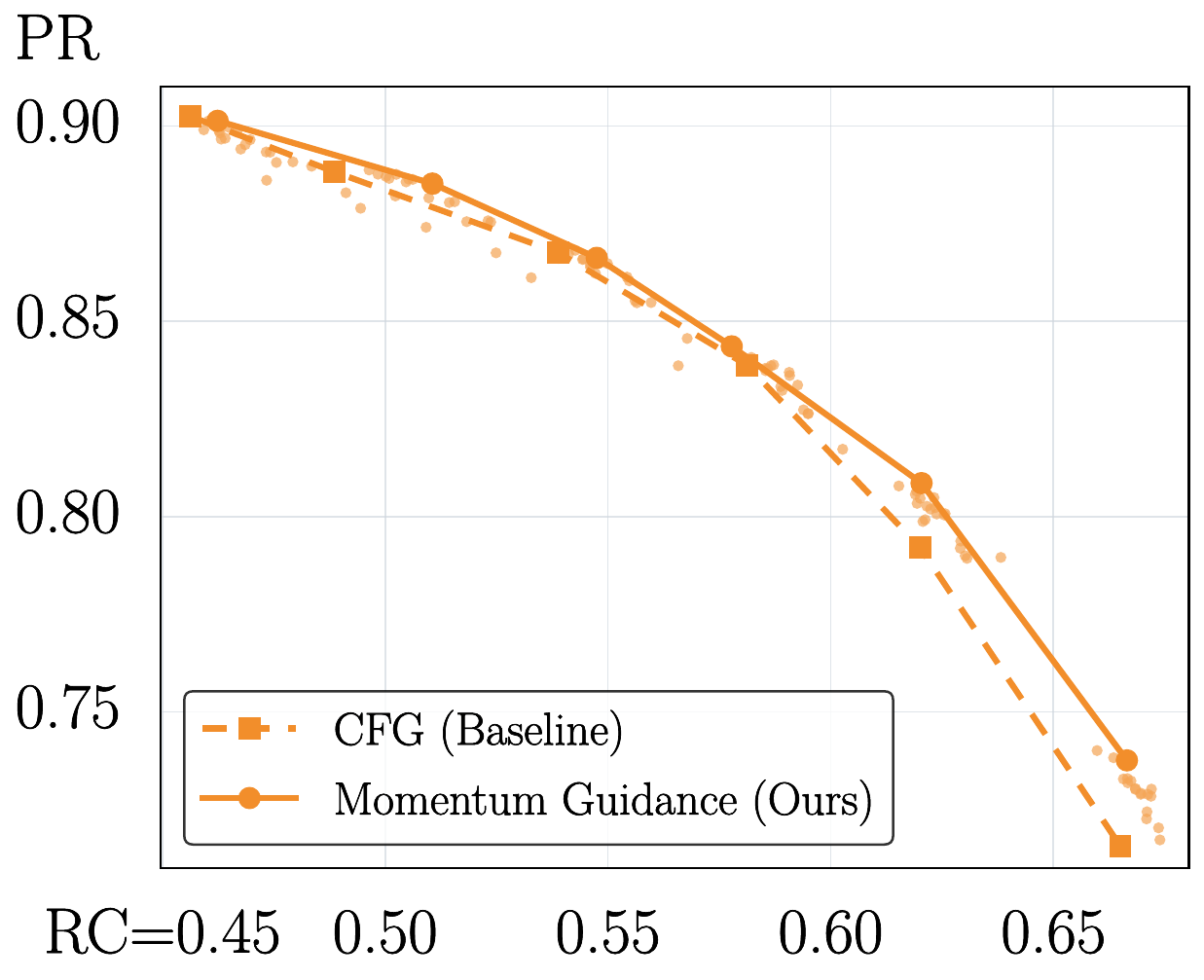}
    \subcaption*{${\textit{NFE}} = 32$}
  \end{subfigure}\hfill
  \begin{subfigure}[t]{0.32\textwidth}
    \centering
    \includegraphics[width=\textwidth]{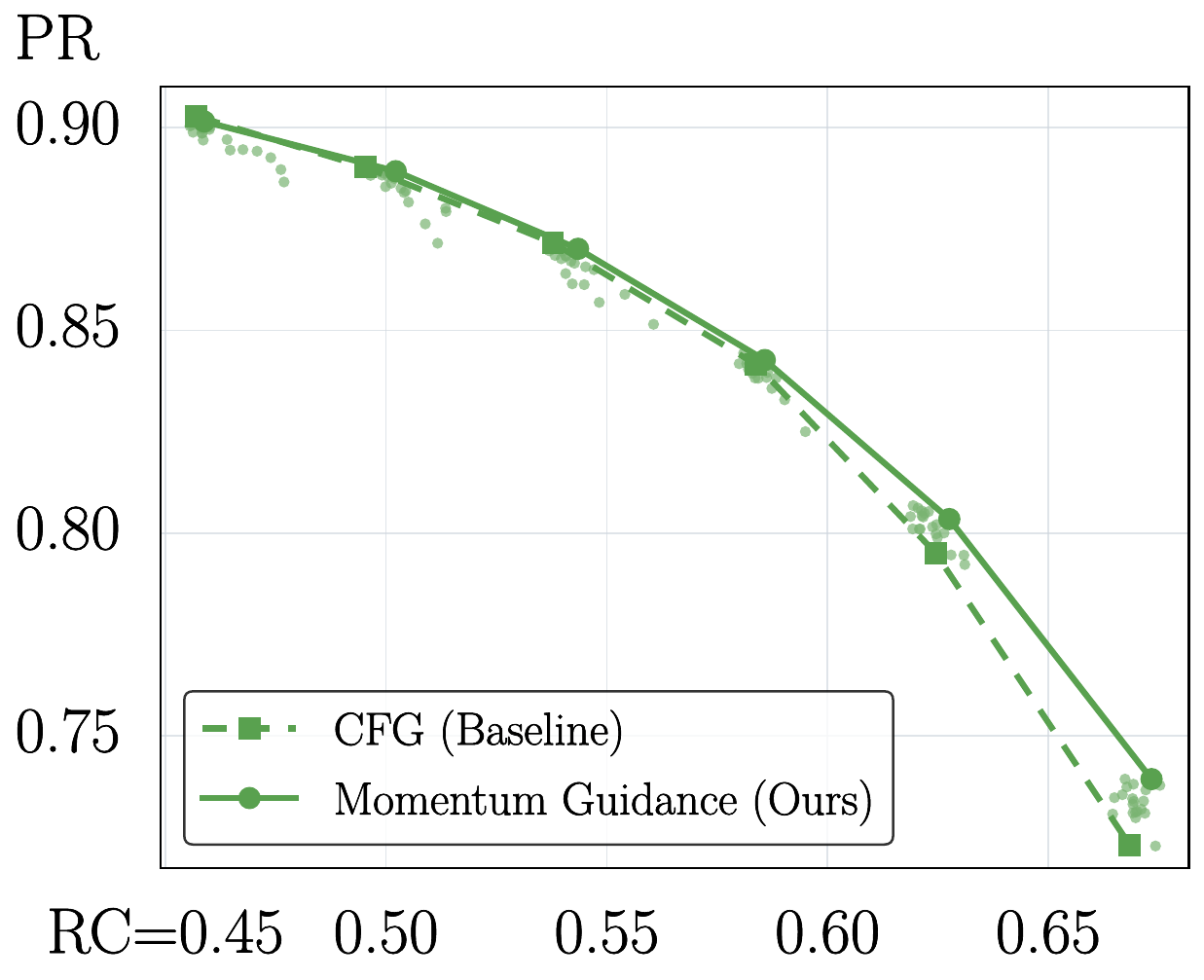}
    \subcaption*{${\textit{NFE}} = 64$}
  \end{subfigure}

  \caption{Ablation over CFG scale and sampling budget on ImageNet-256. \textbf{Top row:} FID as a function of CFG scale for \(\textit{NFE}\!=\!16,32,64\). Solid curves show the best MG configuration for each setting, and shaded bands show the remaining \((\alpha,\beta)\) configurations. \textbf{Bottom row:} Precision--Recall fronts induced by varying the CFG scale. MG consistently lowers FID and shifts the precision--recall trade-off outward relative to vanilla CFG\@.}

  \label{fig:cfg_fid_pr_2x3}
\end{figure*}

\paragraph{Ablation on CFG scale and NFE.}
Figure~\ref{fig:cfg_fid_pr_2x3} studies MG across CFG scales and sampling budgets. For each \((\text{CFG},\text{NFE})\) pair, we report the best MG configuration and use the shaded band to show the range of other \((\alpha,\beta)\) choices. Across all budgets, MG lowers the FID curve relative to vanilla CFG rather than merely shifting the best CFG scale. The improvement is largest at small sampling budgets, especially \(\text{NFE}=16\), where most MG configurations already outperform the baseline. The precision--recall plots show the same effect from another angle: increasing CFG alone improves precision by sacrificing recall, whereas MG expands the frontier, improving precision while better preserving diversity.

\begin{figure*}[t]
  \centering

  \begin{subfigure}[t]{0.32\textwidth}
    \centering
    \includegraphics[width=\textwidth]{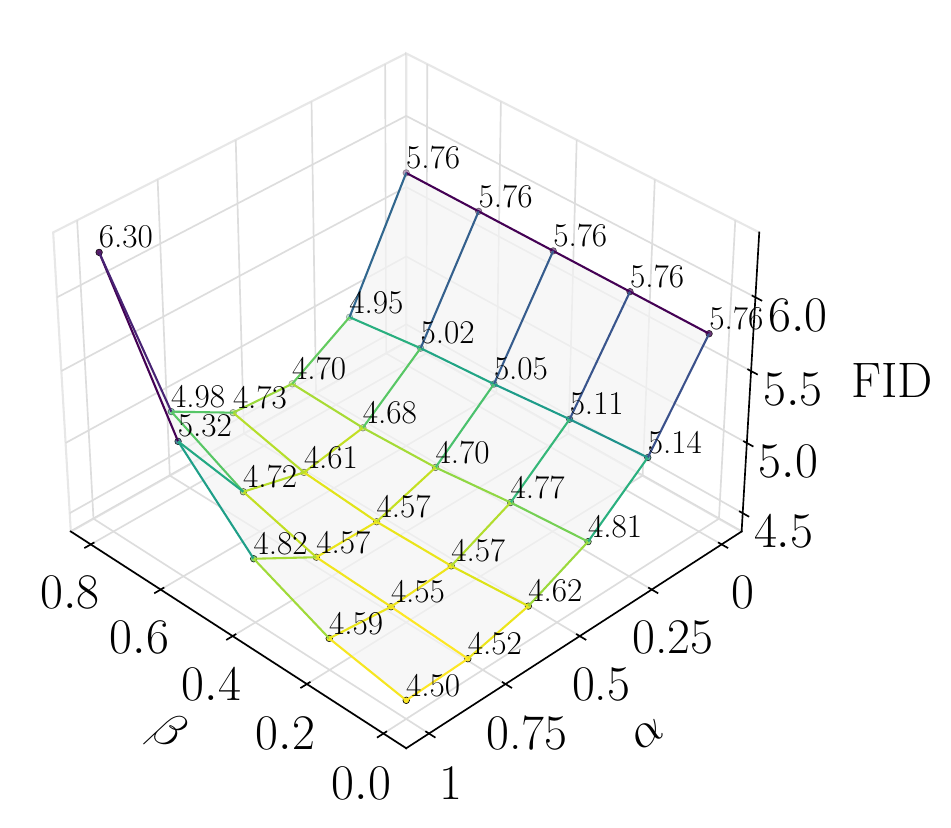}
    \subcaption*{${\textit{NFE}} = 16$}
  \end{subfigure}\hfill
  \begin{subfigure}[t]{0.32\textwidth}
    \centering
    \includegraphics[width=\textwidth]{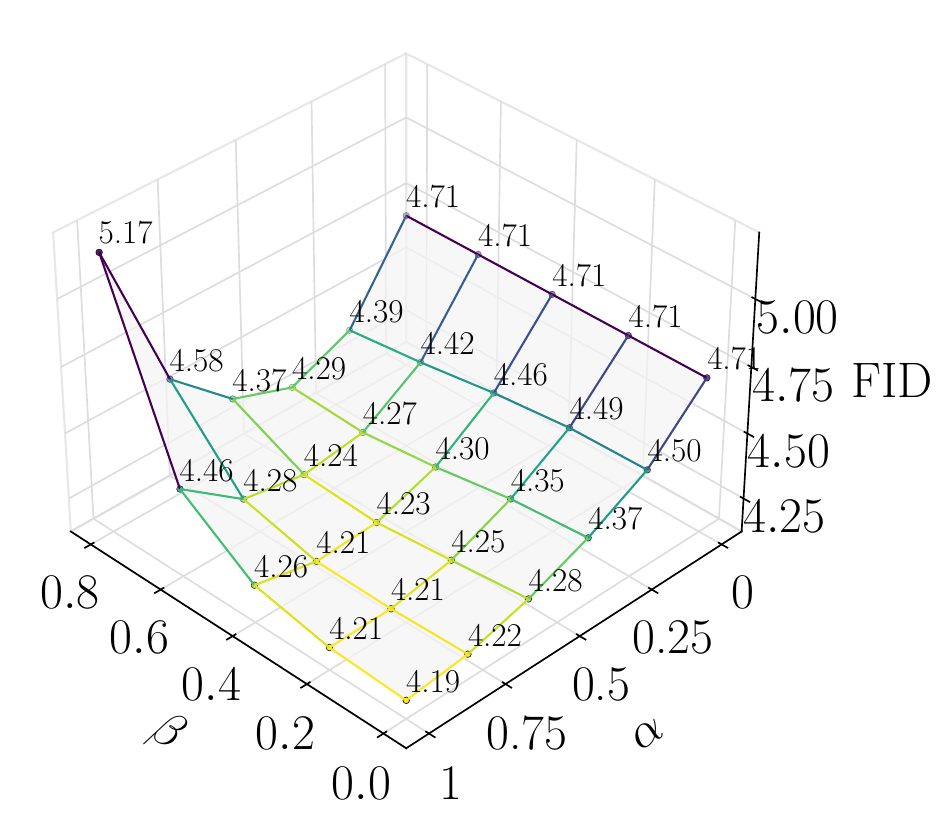}
    \subcaption*{${\textit{NFE}} = 32$}
  \end{subfigure}\hfill
  \begin{subfigure}[t]{0.32\textwidth}
    \centering
    \includegraphics[width=\textwidth]{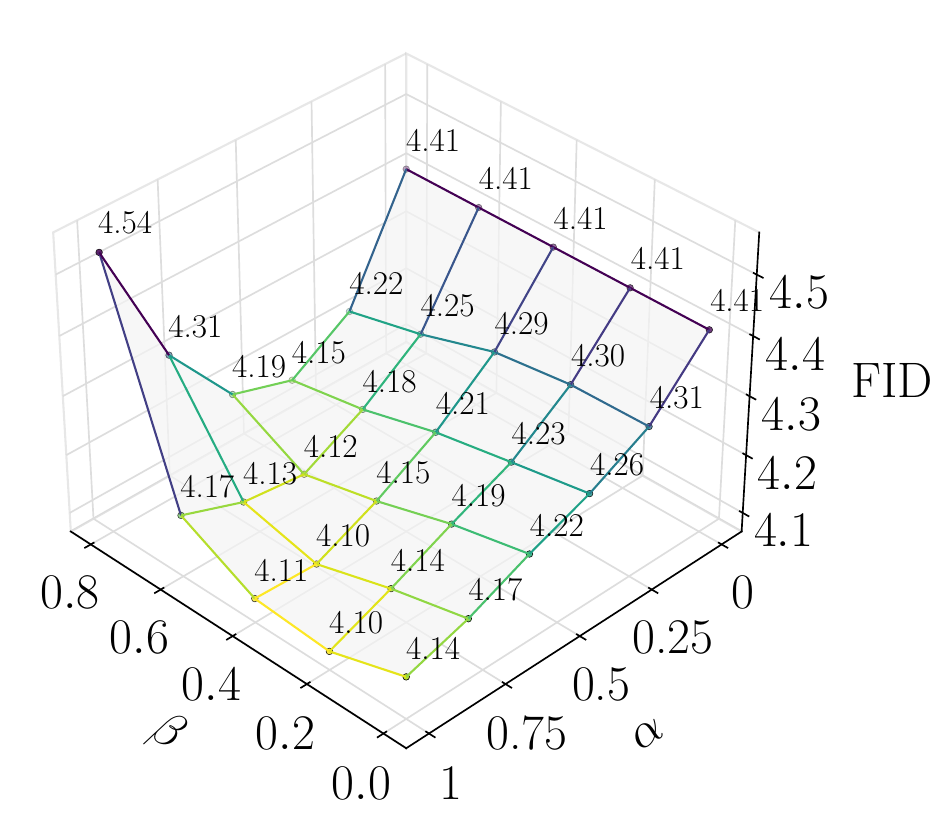}
    \subcaption*{${\textit{NFE}} = 64$}
  \end{subfigure}

  \caption{FID-10K landscape over Momentum Guidance hyperparameters $(\alpha,\beta)$ at $\text{CFG}=1.2$.
Across sampling budgets, nonzero MG strengths consistently improve over the $\alpha=0$ baseline across a wide range of EMA decays, with the best region occurring at moderate $\alpha$ and small-to-medium $\beta$.
}
  \label{fig:3Dsurface}
\end{figure*}

\paragraph{Ablation on $\alpha$ and $\beta$.}
Figure~\ref{fig:3Dsurface} visualizes the FID-10K landscape over the guidance strength $\alpha$ and EMA decay $\beta$ at $\text{CFG}\!=\!1.2$ on ImageNet-256. The $\alpha=0$ edge corresponds to the vanilla CFG baseline. Across \(\text{NFE}=16,32,64\), the same structure appears: increasing $\alpha$ from zero improves FID over a broad region, while excessively large $\alpha$ or overly persistent momentum can over-correct the velocity and degrade quality. The best region typically pairs moderate $\alpha$ with small-to-medium $\beta$, suggesting that strong extrapolation benefits from a shorter momentum memory.

\FloatBarrier

\subsection{Qualitative analysis}

Figure~\ref{fig:cfg1p0cmp} shows a comparison between baseline sampling without CFG and our Momentum Guidance applied to the same SD3 backbone, also without CFG\@. The baseline images often appear blurry and lack coherent structure without the sharpening effect of CFG\@. In contrast, our method produces higher image quality and clearer local structures while retaining the original scene layout. Figure~\ref{fig:sd3difcfg} compares the SD3 baseline and our method across different CFG scales. While the baseline becomes blurry at low CFG and overly saturated at high CFG, our method provides higher image quality, showing that MG reliably improves image quality across a wide range of guidance strengths.

\begin{figure}[!ht]
  \centering
  \includegraphics[width=\linewidth]{images/sd3_difCFG.pdf}
  \caption{Qualitative comparison across CFG scales on SD3\@. MG improves visual detail and structural stability over the baseline across guidance strengths, sharpening low-CFG samples while preserving cleaner textures, balanced contrast, and more stable geometry at stronger CFG scales.}
  \label{fig:sd3difcfg}
\end{figure}

\FloatBarrier

\section{Related Work}

\paragraph{Guidance Methods}
Classifier guidance introduced an inference-time trade-off between mode coverage and sample fidelity by adding classifier gradients to the diffusion score~\citep{dhariwal2021diffusion}. Classifier-free guidance (CFG) removes the external classifier by combining conditional and unconditional model predictions~\citep{ho2022classifier}, and underlies many text-to-image diffusion systems, including GLIDE, Imagen, and latent diffusion models~\citep{nichol2021glide, saharia2022photorealistic, rombach2022high}. Controllable synthesis methods extend the same conditional diffusion interface through structural conditioning and cross-attention manipulation~\citep{zhang2023adding, hertz2022prompt, cao2024controllable}.

Recent variants mainly differ in how they modify the CFG update or construct a reference prediction. Training-free methods include limited-interval guidance, CFG++, and FBG, which respectively schedule, constrain, or adapt the guidance update~\citep{kynkaanniemi2024applying, chung2025cfg, koulischer2025feedback}; attention-based methods such as SAG, PAG, and SEG, which derive degraded or smoothed predictions from the same network~\citep{hong2023improving, ahn2024self, hong2024smoothed}; and direction or coefficient corrections such as APG, ADG, TCFG, and ReCFG~\citep{sadat2025eliminating, jin2025angle, kwon2025tcfg, xia2025rectified}. CADS instead anneals the conditioning signal to recover diversity at high guidance scales~\citep{sadat2023cads}. Auxiliary or training-based approaches change or learn the reference: Autoguidance uses a weaker generator as the reference branch~\citep{karras2024guiding}, while guidance distillation trains a student or lightweight guide to amortize CFG~\citep{meng2023distillation, hsiao2024plug}. Outside generative guidance, recent work has also explored training-free iterative computation, guided adversarial self-play, and mixture-of-experts balancing as complementary mechanisms for improving model behavior~\citep{chen2026training,li2026learning,chen2026phi}. MG is complementary: it obtains the smoother reference from the current ODE trajectory through a velocity EMA, requiring no unconditional branch, auxiliary checkpoint, or extra network evaluation.

\section{Conclusions}

We presented \emph{Momentum Guidance} (MG), an inference-time guidance method that uses the sampler's own ODE trajectory to construct a smoother velocity reference. By maintaining an EMA of past velocities and extrapolating the current velocity away from it, MG produces the sharpening effect of guidance without an unconditional branch, auxiliary checkpoint, or additional network evaluation. Across ImageNet-256, large-scale text-to-image models such as SD3 and FLUX.1-dev, and text-to-video generation with HunyuanVideo, MG consistently improves sample fidelity and structural detail. On ImageNet, it further expands the precision--recall Pareto frontier beyond what is achieved by tuning CFG alone. These results show that trajectory history is a useful and underexplored source of guidance for flow-based generative sampling, offering a simple complement to CFG and related guidance schemes.

\section*{Acknowledgements}
This work was supported in part by the Institute for Foundations of Machine Learning (IFML). The authors acknowledge the Texas Advanced Computing Center (TACC) at The University of Texas at Austin for providing computational resources that have contributed to the research results reported within this paper.

\bibliographystyle{splncs04}
\bibliography{main}

\clearpage
\section*{Appendix}
\setcounter{section}{0}
\renewcommand{\thesection}{\Alph{section}}
\renewcommand{\theHsection}{appendix.\Alph{section}}
\renewcommand{\theHsubsection}{appendix.\Alph{section}.\arabic{subsection}}
\makeatletter
\providecommand*{\theHALG@line}{}
\renewcommand*{\theHALG@line}{supp.\arabic{ALG@line}}
\makeatother

\section{Momentum Guidance with CFG}

\subsection{CFG-Adjusted Velocity}

Momentum Guidance can be applied either to the conditional sampler alone or on top of classifier-free guidance (CFG).
When CFG is enabled, we simply replace the model velocity in Algorithm~\ref{alg:momentum-guidance} with the CFG-adjusted velocity.

Let $v_c$ and $v_u$ denote the conditional and unconditional velocities:
\[
v_c := \vv_\theta(\vx,t,c),
\qquad
v_u := \vv_\theta(\vx,t,\emptyset).
\]
The CFG-augmented velocity is
\[
\vv_\theta^{\mathrm{CFG}}(\vx,t,c;\omega)
\triangleq
\begin{cases}
v_c, & \omega = 1,\\[0.4ex]
\omega\, v_c + (1-\omega)\, v_u, & \omega > 1,
\end{cases}
\]
where $\omega \ge 1$ denotes the CFG scale, and $\omega=1$ recovers conditional sampling without CFG\@.
Algorithm~\ref{alg:momentum-guidance-cfg} applies MG to this redefined velocity.
The same principle could also be used with other reference branches, such as Autoguidance~\citep{karras2024guiding}, but a systematic study of these variants is outside the scope of this paper.

\begin{algorithm}[H]
\caption{\textbf{Momentum Guidance with CFG}}
\label{alg:momentum-guidance-cfg}
\begin{algorithmic}[1]
\Require Conditional flow model $\vv_{\theta}(\cdot,t,c)$; unconditional branch $\vv_{\theta}(\cdot,t,\emptyset)$; condition $c$; time grid $\{t_i\}_{i=0}^N$; EMA decay $\beta\in[0,1)$; MG strength $\alpha \geq 0$; CFG scale $\omega \ge 1$; CFG velocity $\vv_\theta^{\mathrm{CFG}}(\vx,t,c;\omega)$
\State Sample $\mZ_{t_0}\sim\mathcal{N}(0,\mI)$
\State Initialize momentum $\vm_{t_0} \gets \vv_\theta^{\mathrm{CFG}}(\mZ_{t_0},t_0,c;\omega)$
\For{$i=0$ to $N-1$}
  \State $\Delta t\gets t_{i+1}-t_i$
  \vspace{0.3ex}
  \State $\vv_{t_i} \gets \vv_\theta^{\mathrm{CFG}}(\mZ_{t_i},t_i,c;\omega)$
  \vspace{0.3ex}
  \State
  $
    \mZ_{t_{i+1}} \gets \mZ_{t_i}
    + \Delta t \Big[\,\vv_{t_i} +
    \colorbox{orange!12}{\text{$\alpha(\vv_{t_i}-\,\vm_{t_{i}})$}}
    \Big]
  $
  \State  \colorbox{orange!12}{$\vm_{t_{i+1}}\gets (1-\beta)\,\vv_{t_i}+\beta\,\vm_{t_i}$}  \Comment{EMA}
  \vspace{0.3ex}
\EndFor
\State \Return $\mZ_{t_N}$
\end{algorithmic}
\end{algorithm}

\subsection{CFG Interval}
Guidance interval~\citep{kynkaanniemi2024applying} applies CFG only over a selected range of flow times.
This schedule reduces the diversity loss often introduced by strong CFG, since the unconditional branch is disabled outside the chosen interval.
We use it as an additional comparison: the main experiments apply CFG at every timestep, while Table~\ref{tab:cfg_vs_ours} already shows that MG improves both conditional sampling without CFG and standard CFG\@.

Table~\ref{tab:ema_euler_cfg_interval_appendix} evaluates MG on top of the interval schedule $t\in[0.125,1]$, with MG itself applied over the full flow interval.
At $\omega\!=\!1.4$, MG gives a clear gain over the CFG-interval baseline and reaches $\text{FID}\!=\!1.553$ with only 16 NFEs.
At $\omega\!=\!1.6$, MG improves the 16- and 32-step results and remains competitive at 64 steps, roughly matching FID while improving IS and recall.

\begin{table*}[t]
    \centering
    \caption{CFG-interval results on ImageNet-256~\citep{kynkaanniemi2024applying}. CFG is applied only on \(t\in[0.125,1]\), while MG is applied over the full flow interval. We compare standard CFG, CFG interval, and CFG interval combined with MG using the selected \((\alpha,\beta)\).}
    \label{tab:ema_euler_cfg_interval_appendix}
    \footnotesize
    \begin{tabular}{cccccccc}
        \toprule
        CFG $\omega$ & NFE & Method & $(\alpha,\beta)$ & FID-50K $\downarrow$ & IS $\uparrow$ & Precision $\uparrow$ & Recall $\uparrow$ \\
        \midrule
        \multirow{9}{*}[-1.1ex]{1.4}
            & \multirow{3}{*}{16}
                & CFG & $-$
                    & 2.380 & 275.06 & 0.827 & 0.568 \\
            &   & CFG interval & $-$
                    & 2.352 & 249.85 & 0.791 & 0.612 \\
            &   & \cellcolor{lightpink}CFG interval + MG
                    & \cellcolor{lightpink}$(0.8, 0.0)$
                    & \cellcolor{lightpink}\textbf{1.553}
                    & \cellcolor{lightpink}268.03
                    & \cellcolor{lightpink}0.799
                    & \cellcolor{lightpink}\textbf{0.636} \\
        \cmidrule(lr){2-8}
            & \multirow{3}{*}{32}
                & CFG & $-$
                    & 2.039 & 293.03 & 0.839 & 0.581 \\
            &   & CFG interval & $-$
                    & 1.642 & 264.75 & 0.800 & 0.626 \\
            &   & \cellcolor{lightpink}CFG interval + MG
                    & \cellcolor{lightpink}$(0.6, 0.2)$
                    & \cellcolor{lightpink}\textbf{1.408}
                    & \cellcolor{lightpink}274.99
                    & \cellcolor{lightpink}0.801
                    & \cellcolor{lightpink}\textbf{0.633} \\
        \cmidrule(lr){2-8}
            & \multirow{3}{*}{64}
                & CFG & $-$
                    & 2.028 & 301.29 & 0.842 & 0.584 \\
            &   & CFG interval & $-$
                    & 1.462 & 271.34 & 0.803 & 0.625 \\
            &   & \cellcolor{lightpink}CFG interval + MG
                    & \cellcolor{lightpink}$(0.6, 0.4)$
                    & \cellcolor{lightpink}\textbf{1.380}
                    & \cellcolor{lightpink}277.73
                    & \cellcolor{lightpink}0.802
                    & \cellcolor{lightpink}\textbf{0.630} \\
        \midrule
        \multirow{9}{*}[-1.1ex]{1.6}
            & \multirow{3}{*}{16}
                & CFG & $-$
                    & 3.128 & 325.55 & 0.862 & 0.524 \\
            &   & CFG interval & $-$
                    & 1.993 & 291.80 & 0.819 & 0.594 \\
            &   & \cellcolor{lightpink}CFG interval + MG
                    & \cellcolor{lightpink}$(0.6, 0.2)$
                    & \cellcolor{lightpink}\textbf{1.638}
                    & \cellcolor{lightpink}306.57
                    & \cellcolor{lightpink}0.822
                    & \cellcolor{lightpink}\textbf{0.611} \\
        \cmidrule(lr){2-8}
            & \multirow{3}{*}{32}
                & CFG & $-$
                    & 3.170 & 340.88 & 0.868 & 0.539 \\
            &   & CFG interval & $-$
                    & 1.639 & 305.82 & 0.822 & 0.605 \\
            &   & \cellcolor{lightpink}CFG interval + MG
                    & \cellcolor{lightpink}$(0.2, 0.8)$
                    & \cellcolor{lightpink}\textbf{1.581}
                    & \cellcolor{lightpink}308.74
                    & \cellcolor{lightpink}0.822
                    & \cellcolor{lightpink}\textbf{0.613} \\
        \cmidrule(lr){2-8}
            & \multirow{3}{*}{64}
                & CFG & $-$
                    & 3.342 & 348.87 & 0.872 & 0.538 \\
            &   & CFG interval & $-$
                    & \textbf{1.612} & 311.17 & 0.824 & 0.608 \\
            &   & \cellcolor{lightpink}CFG interval + MG
                    & \cellcolor{lightpink}$(0.2, 0.4)$
                    & \cellcolor{lightpink}1.618
                    & \cellcolor{lightpink}314.33
                    & \cellcolor{lightpink}0.824
                    & \cellcolor{lightpink}\textbf{0.613} \\
        \bottomrule
    \end{tabular}
\end{table*}

\begin{figure*}[t]
    \captionsetup[subfigure]{font=normalsize}
    \centering
    \begin{subfigure}[b]{\linewidth}
        \centering
        \includegraphics[width=\linewidth]{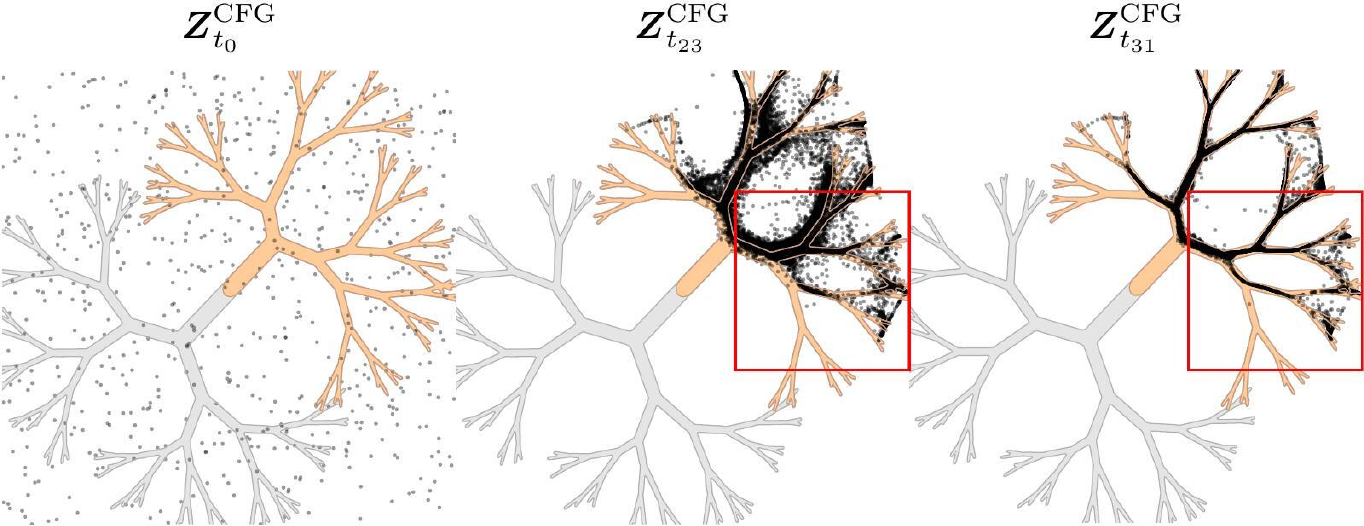}
        \caption{CFG baseline trajectories.}
        \label{fig:toy1}
    \end{subfigure}
    \hfill
    \begin{subfigure}[b]{\linewidth}
        \centering
        \includegraphics[width=\linewidth]{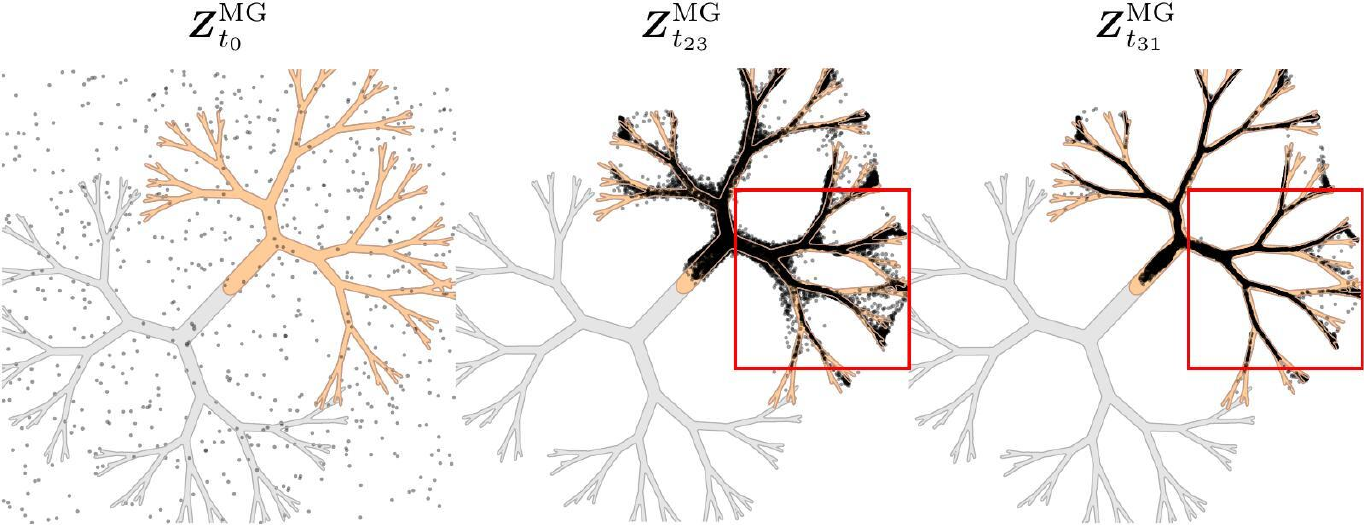}
        \caption{MG trajectories.}
        \label{fig:toy2}
    \end{subfigure}
    \hfill
    \begin{subfigure}[b]{\linewidth}
        \centering
        \includegraphics[width=\linewidth]{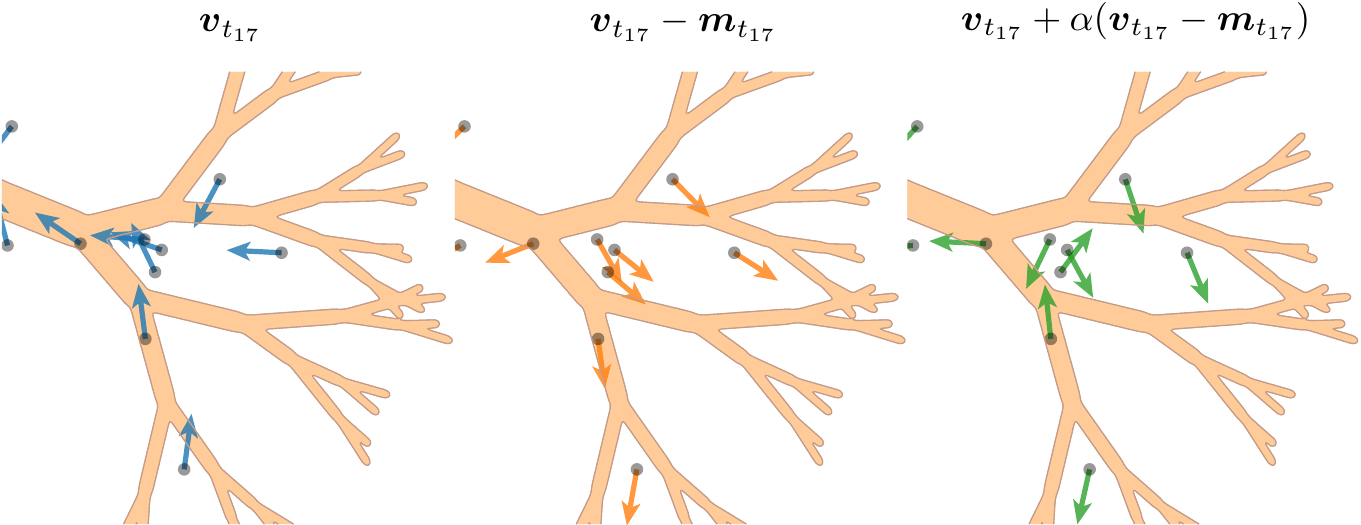}
        \caption{Velocity-field diagnostic.}
        \label{fig:toy3}
    \end{subfigure}

    \caption{\textbf{2D Gaussian-mixture toy.} We compare a CFG Euler sampler with MG on a tree-shaped binary mixture. MG preserves a wider spread of particles along the target branch. The velocity-field diagnostic visualizes how the extrapolation direction \(\boldsymbol v_t-\boldsymbol m_t\) counters the pull toward the conditional mode center.}
    \label{fig:toy}
\end{figure*}

\begin{figure*}[t]
    \centering
    \begin{subfigure}[b]{\linewidth}
        \centering
        \includegraphics[width=\linewidth]{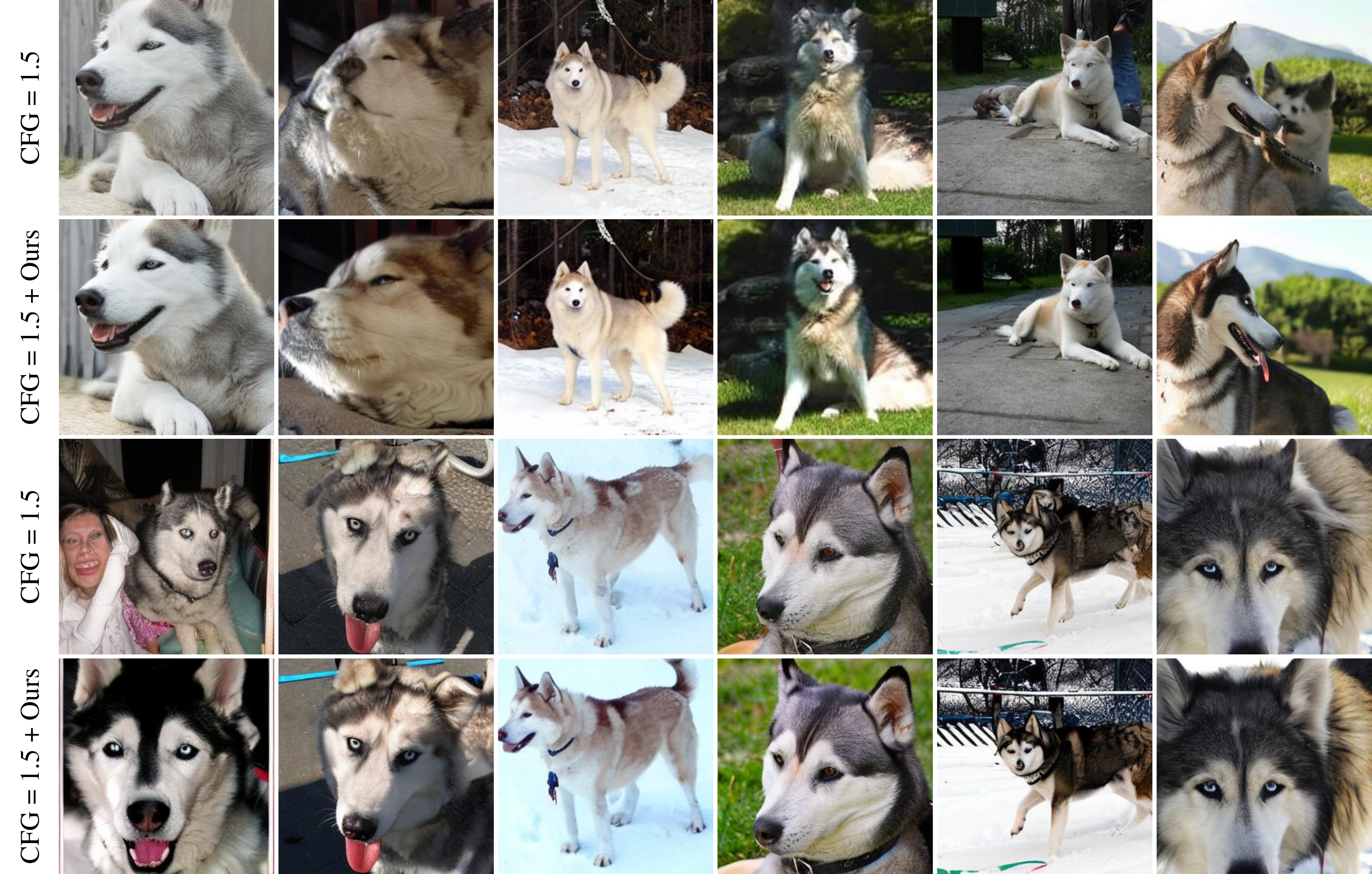}
        \caption{Class 248: \textit{Eskimo dog, husky}.}
        \label{fig:imagenet-class248}
    \end{subfigure}
    \hfill
    \begin{subfigure}[b]{\linewidth}
        \centering
        \includegraphics[width=\linewidth]{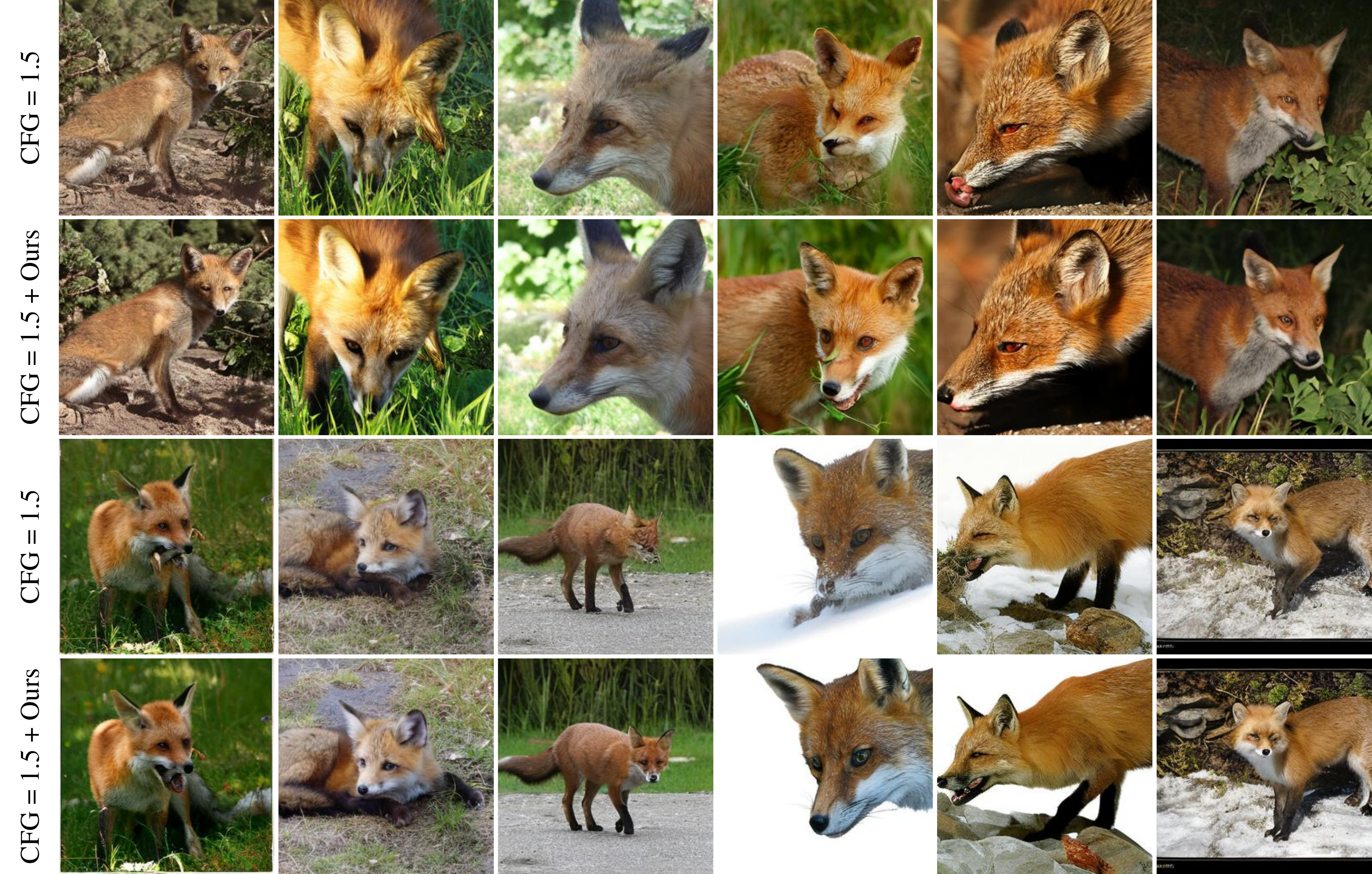}
        \caption{Class 277: \textit{red fox, Vulpes vulpes}.}
        \label{fig:imagenet-class278}
    \end{subfigure}
    \caption{\textbf{ImageNet-256 qualitative comparison.} Samples use 32 Euler steps and CFG \(=1.5\). MG is applied to the same conditional model with \(\alpha=1.0\), \(\beta=0.6\), and interval \(t\in[0.1,0.7]\), improving local structure while preserving the class identity and global composition.}
    \label{fig:imagenet-class248-class278}
\end{figure*}
\begin{figure*}[t]
    \centering
    \begin{subfigure}[b]{\linewidth}
        \centering
        \includegraphics[width=\linewidth]{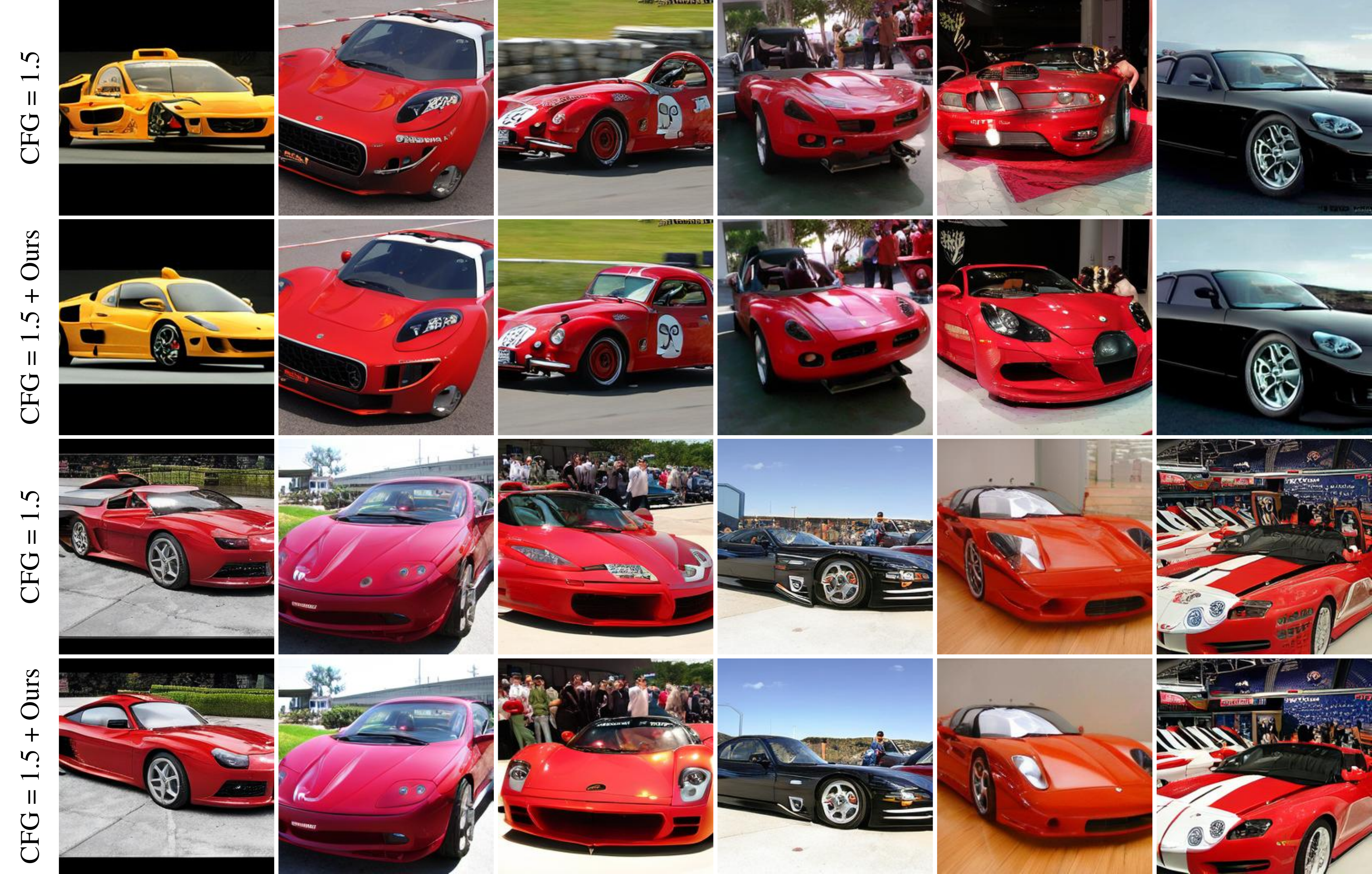}
        \caption{Class 817: \textit{sports car, sport car}.}
        \label{fig:imagenet-class817}
    \end{subfigure}
    \hfill
    \begin{subfigure}[b]{\linewidth}
        \centering
        \includegraphics[width=\linewidth]{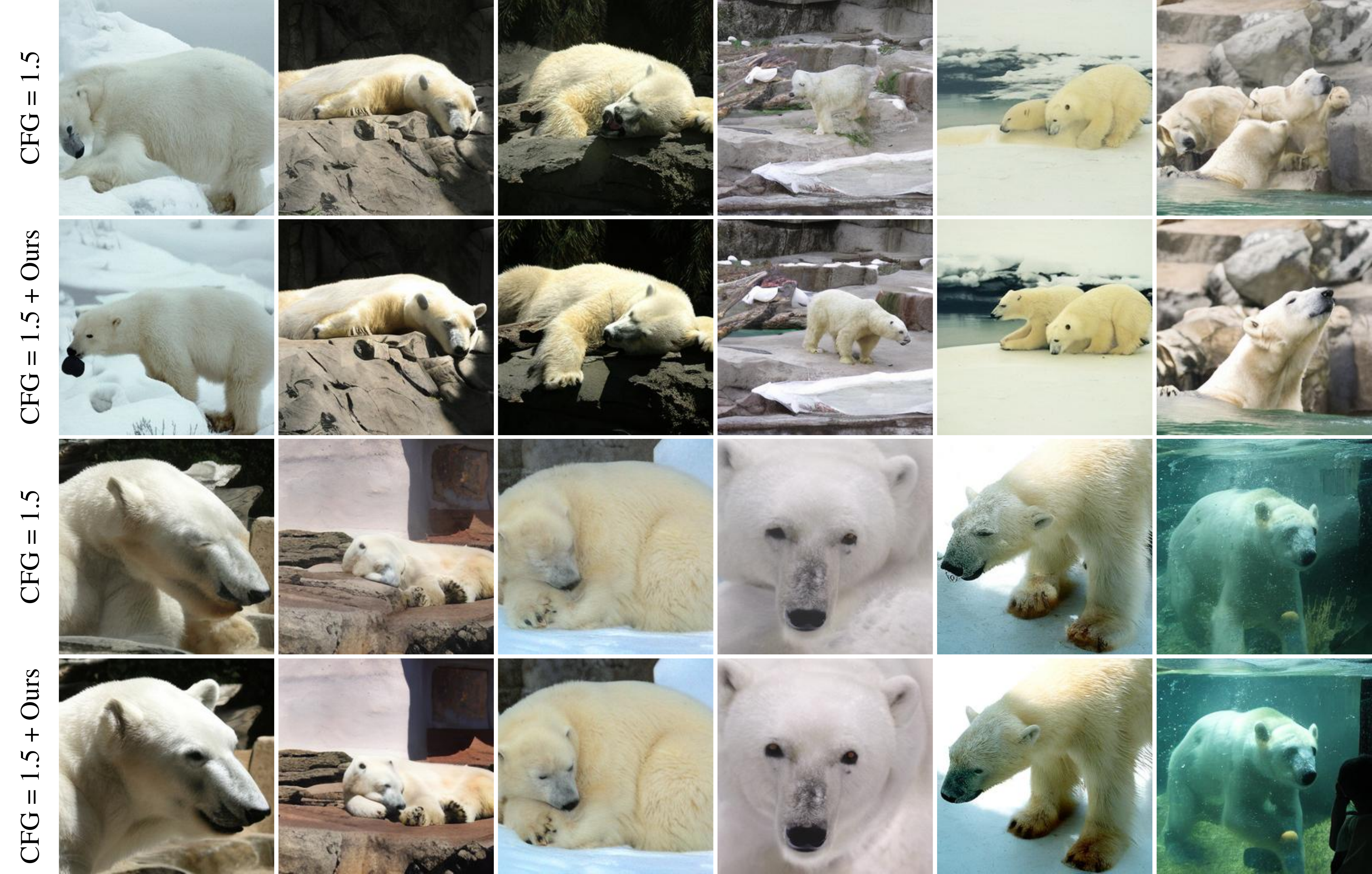}
        \caption{Class 296: \textit{ice bear, polar bear}.}
        \label{fig:imagenet-class296}
    \end{subfigure}
    \caption{\textbf{ImageNet-256 qualitative comparison.} Samples use 32 Euler steps and CFG \(=1.5\). MG reduces common CFG artifacts such as distorted object parts, blurred local regions, and unstable geometry, while maintaining the overall scene layout.}
    \label{fig:imagenet-class817-class296}
\end{figure*}

\section{Additional Implementation Details}

\subsection{ImageNet-256 Setup}

We evaluate Momentum Guidance using the improved DiT-XL checkpoint from the official Rectified Flow codebase~\citep{let2025Liu}.
The model is trained for \(400{,}000\) steps with a global batch size of \(2048\), EMA decay \(0.9999\), learning rate \(2\times 10^{-4}\), and the logit-normal time-sampling distribution introduced by SD3~\citep{esser2024scaling}.

FID, IS, and precision--recall are computed following the evaluation protocol of~\citep{dhariwal2021diffusion}.
For each CFG scale and NFE budget, we sweep \((\alpha,\beta)\) on a grid with spacing \(0.2\), select the configuration by FID-10K, and report the corresponding FID-50K, IS, precision, and recall.
All configurations share the same initial noise batch, so metric differences reflect the sampling rule rather than random variation in the initial latent.

Although Table~\ref{tab:cfg_vs_ours} reports FID-selected configurations, Figure~\ref{fig:cfg_fid_pr_2x3} shows the full sweep through the shaded bands.
Most nonzero MG settings improve over the baseline across a broad range, indicating that the method is not sensitive to precise hyperparameter tuning.
In practice, a small sweep is still useful for obtaining the strongest result under a given sampler and guidance scale.

\subsection{Optional Normalization and Unbiased EMA}
\label{app:mg-normalization-unbiased}
Beyond the basic EMA update, we consider two small refinements to the velocity momentum.
The basic update initializes the EMA with the first velocity, \(\tilde{\vm}_{t_0}=\vv_{t_0}\).
The debiased variant instead uses a zero-initialized accumulator and applies the standard EMA correction
\[
\vm_{t_i} = \frac{\tilde{\vm}_{t_i}}{1-\beta^{s_i}},
\]
where $s_i$ is the number of EMA updates up to time $t_i$.
We also test a per-sample normalization that matches the \(\ell_2\)-norm of the momentum to that of the current velocity,
\[
\vm_{t_i} \leftarrow
\frac{\|\vv_{t_i}\|_2}{\|\vm_{t_i}\|_2 + \varepsilon}\,\vm_{t_i}.
\]

The debiasing correction prevents a zero-initialized EMA from being underestimated during the earliest flow steps, while normalization removes trivial scale differences between \(\vv_{t_i}\) and the EMA reference.
Neither refinement changes the number of model evaluations or the form of the MG update.

\begin{figure*}[t]
    \centering
    \begin{subfigure}{\textwidth}
        \centering
        \includegraphics[width=\textwidth]{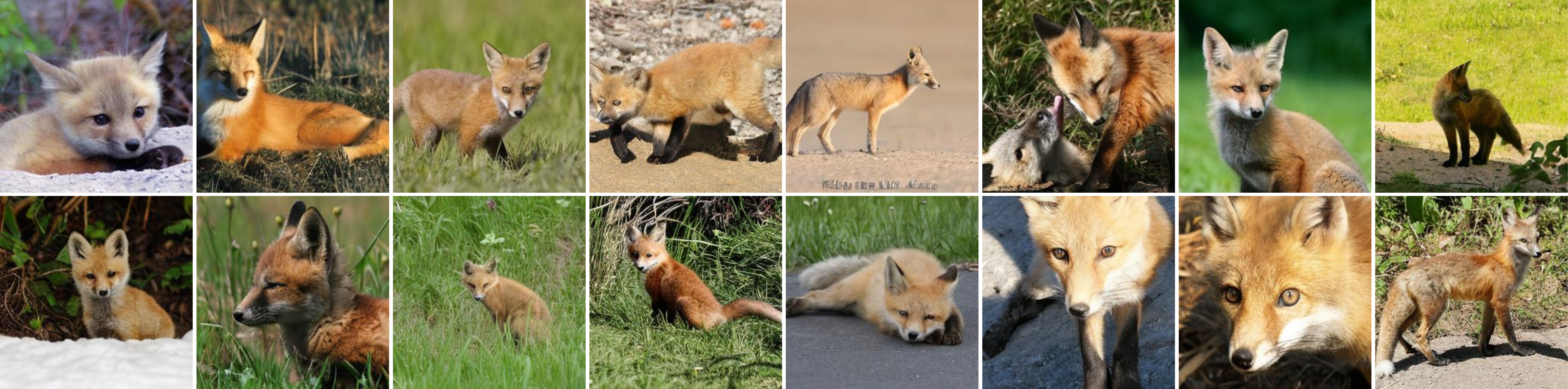}
        \caption{No normalization or debiasing.}
        \label{fig:v_a}
    \end{subfigure}\\[0.4em]
    \begin{subfigure}{\textwidth}
        \centering
        \includegraphics[width=\textwidth]{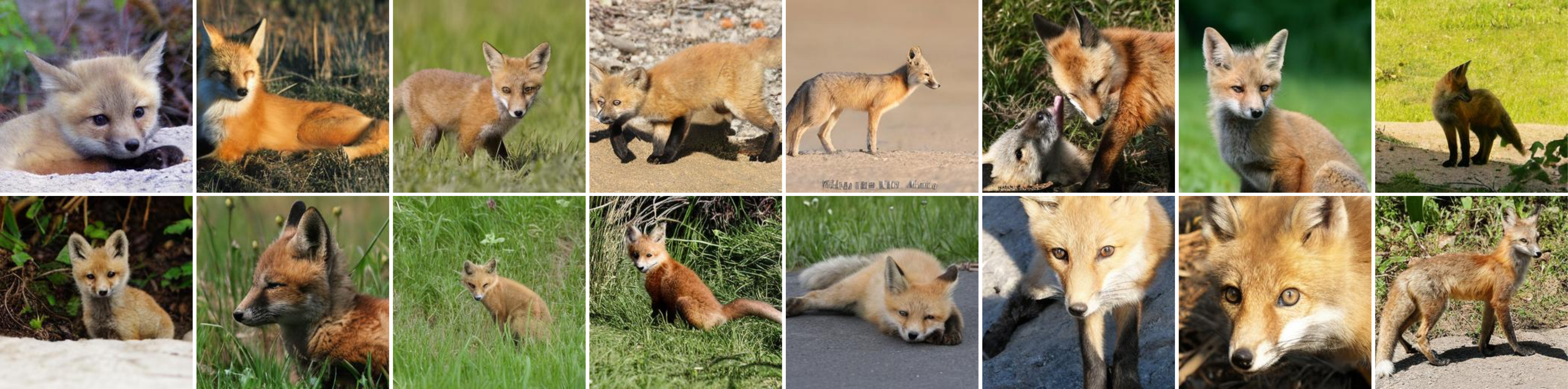}
        \caption{Normalization only.}
        \label{fig:v_b}
    \end{subfigure}\\[0.4em]
    \begin{subfigure}{\textwidth}
        \centering
        \includegraphics[width=\textwidth]{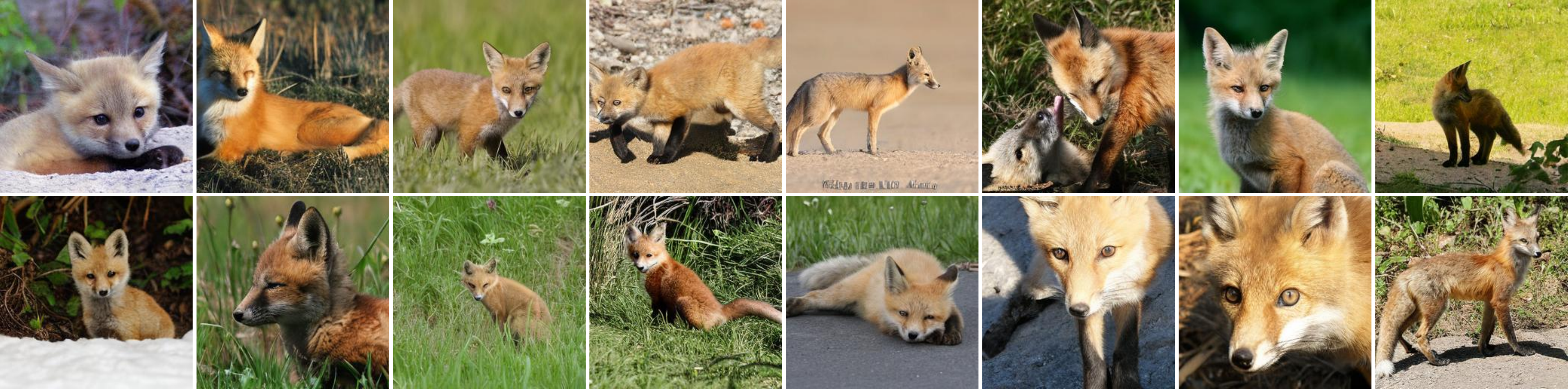}
        \caption{Unbiased EMA only.}
        \label{fig:v_c}
    \end{subfigure}\\[0.4em]
    \begin{subfigure}{\textwidth}
        \centering
        \includegraphics[width=\textwidth]{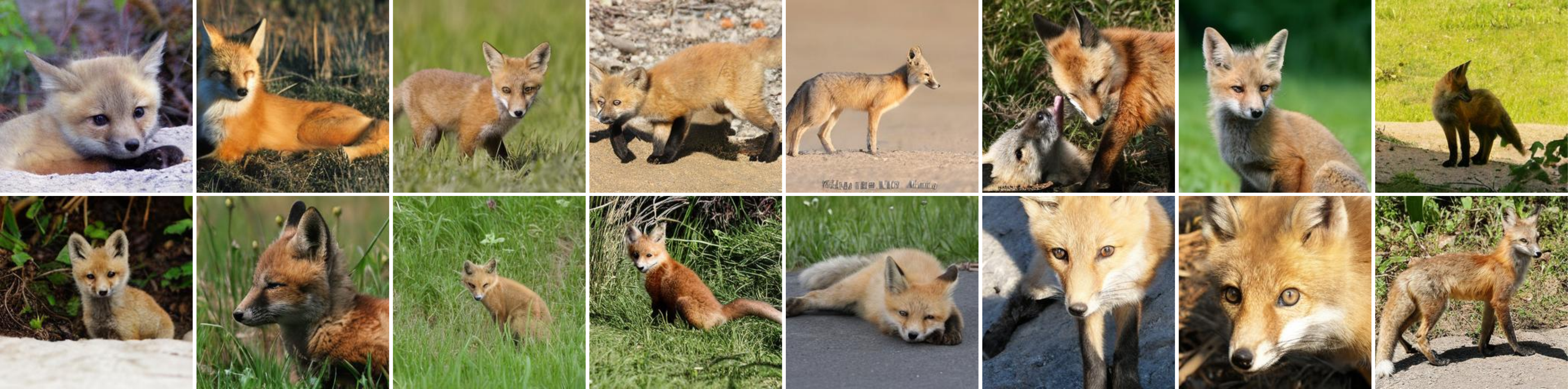}
        \caption{Normalization and unbiased EMA.}
        \label{fig:v_d}
    \end{subfigure}
    \caption{\textbf{Effect of optional normalization and unbiased EMA.} Samples are generated with MG \((\alpha=1.0,\beta=0.6)\) at CFG \(=1.5\) on ImageNet class 278 (kit fox). The four rows toggle normalization and EMA debiasing. Visual differences are minor, indicating that these refinements are implementation details rather than the main source of MG's gains.}
    \label{fig:cmp_norm_unbiased}
\end{figure*}

Figure~\ref{fig:cmp_norm_unbiased} compares four variants obtained by toggling normalization and debiasing at \(\text{CFG}=1.5\) and \(\text{NFE}=32\).
The resulting samples are visually very similar.
Quantitatively, normalization and debiasing give small FID improvements in most settings, but these gains are minor compared with the improvement from MG itself.
We therefore view them as implementation refinements rather than the source of the method's benefit.

\section{Additional Experiment Results}
\paragraph{2D Gaussian Mixture Toy.}
We first use the tree-shaped 2D Gaussian-mixture dataset from~\citep{karras2024guiding} to illustrate the particle-level effect of MG\@.
The binary mixture contains an orange class on the upper-right branch and a gray class on the lower-left branch.
Figure~\ref{fig:toy} compares 32-step Euler trajectories under CFG and MG\@.
Standard CFG pulls particles toward the conditional mode center, while MG preserves a broader spread along the target branch.
The velocity-field diagnostic at \(t_{17}\) shows why: the extrapolation direction \(\boldsymbol v_t-\boldsymbol m_t\) points away from the EMA-smoothed reference and counteracts part of the collapse induced by the current conditional velocity.

\begin{figure*}[t]
  \centering

  \begin{subfigure}[t]{0.333\textwidth}
    \centering
    \includegraphics[width=\textwidth]{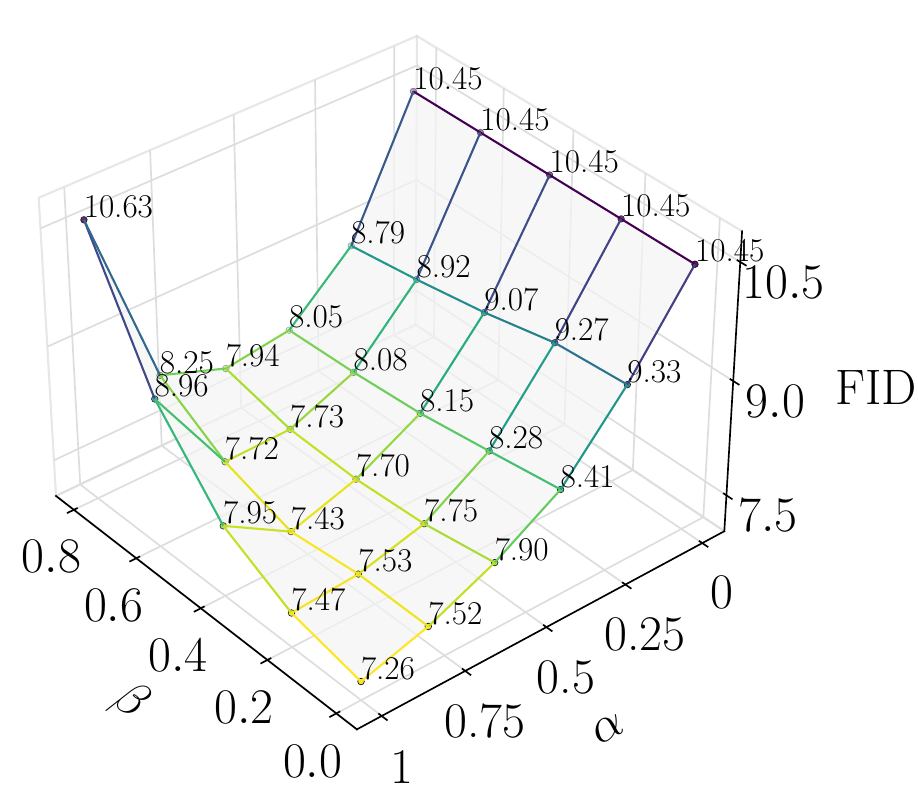}
    \subcaption*{${\textit{NFE}} = 16$}
  \end{subfigure}\hfill
  \begin{subfigure}[t]{0.333\textwidth}
    \centering
    \includegraphics[width=\textwidth]{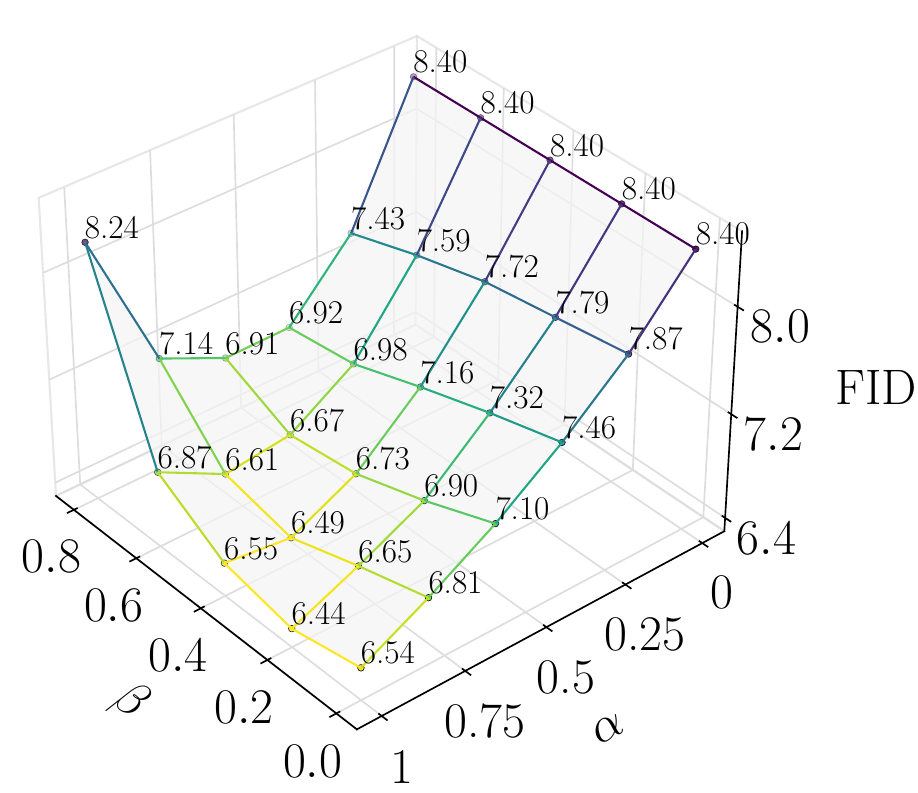}
    \subcaption*{${\textit{NFE}} = 32$}
  \end{subfigure}\hfill
  \begin{subfigure}[t]{0.333\textwidth}
    \centering
    \includegraphics[width=\textwidth]{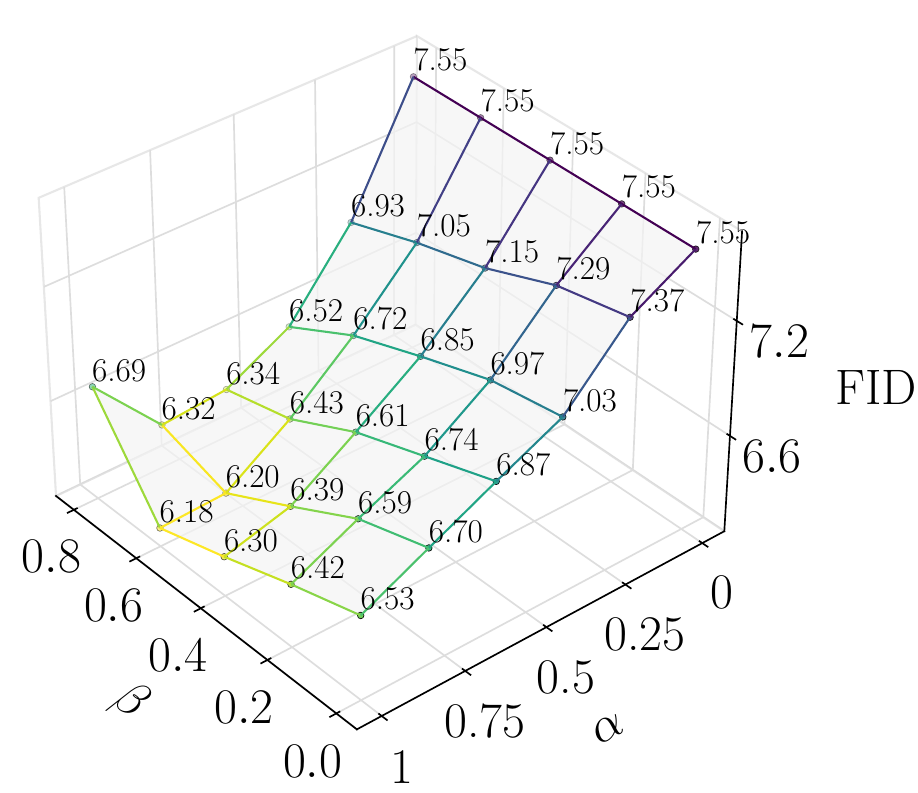}
    \subcaption*{${\textit{NFE}} = 64$}
  \end{subfigure}

  \caption{
FID-10K landscape over MG hyperparameters \((\alpha,\beta)\) without CFG\@.
Across sampling budgets, a broad region of nonzero \(\alpha\) improves over the \(\alpha=0\) baseline, showing that trajectory-based guidance is effective even without CFG\@.
}
  \label{fig:3Dsurface-CFG1p0}
\end{figure*}

\begin{figure*}[t]
  \centering

  \begin{subfigure}[t]{0.333\textwidth}
    \centering
    \includegraphics[width=\textwidth]{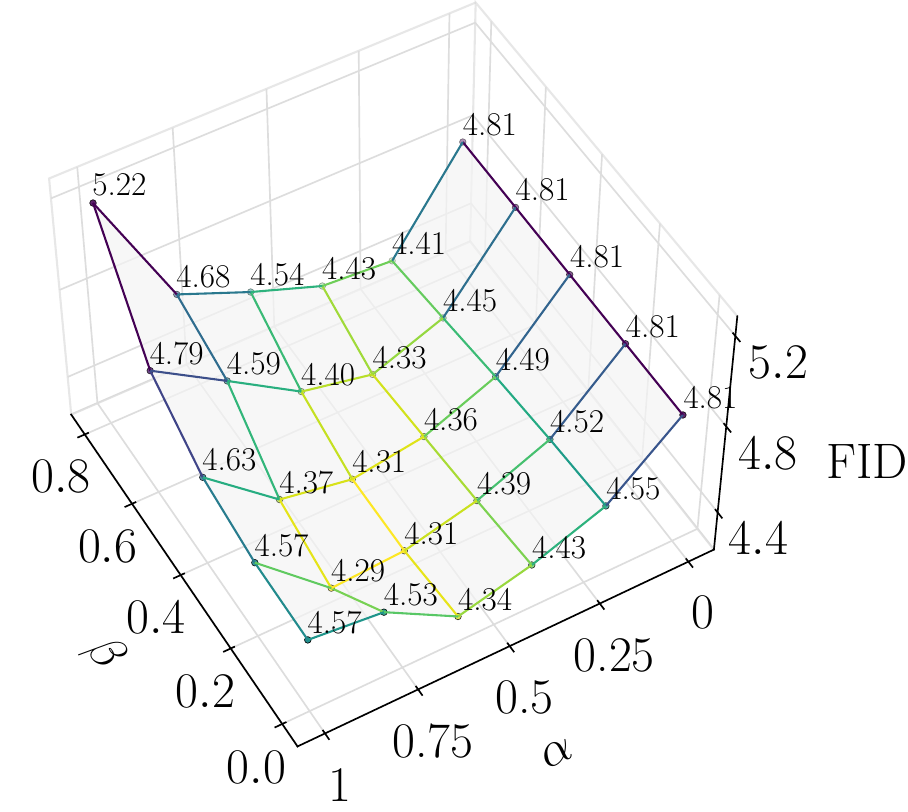}
    \subcaption*{${\textit{NFE}} = 16$}
  \end{subfigure}\hfill
  \begin{subfigure}[t]{0.333\textwidth}
    \centering
    \includegraphics[width=\textwidth]{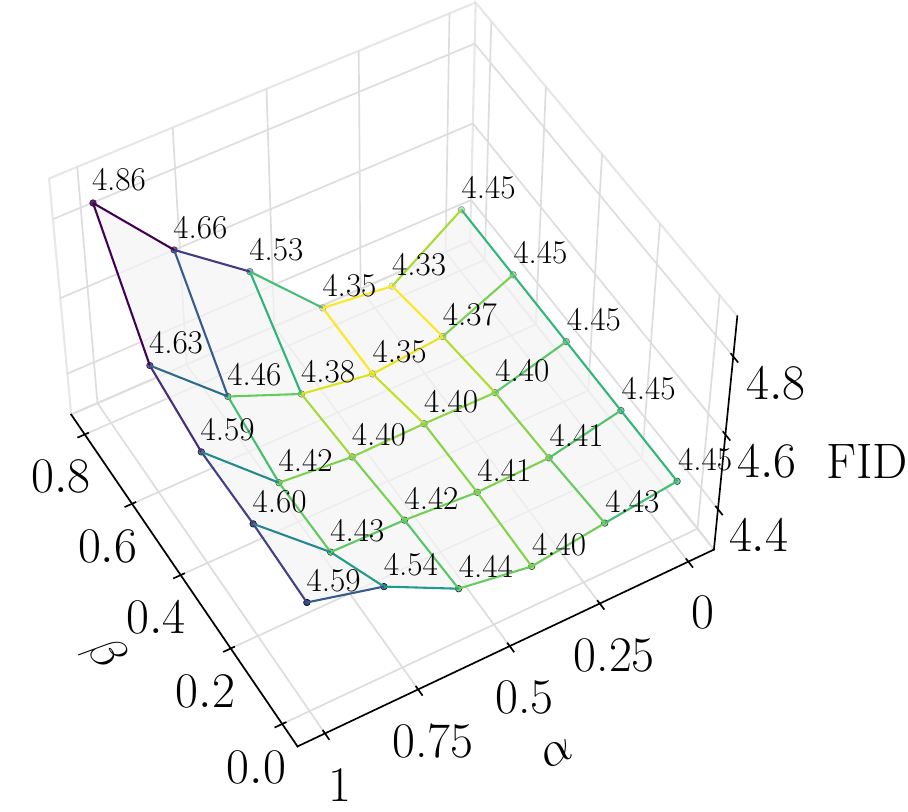}
    \subcaption*{${\textit{NFE}} = 32$}
  \end{subfigure}\hfill
  \begin{subfigure}[t]{0.333\textwidth}
    \centering
    \includegraphics[width=\textwidth]{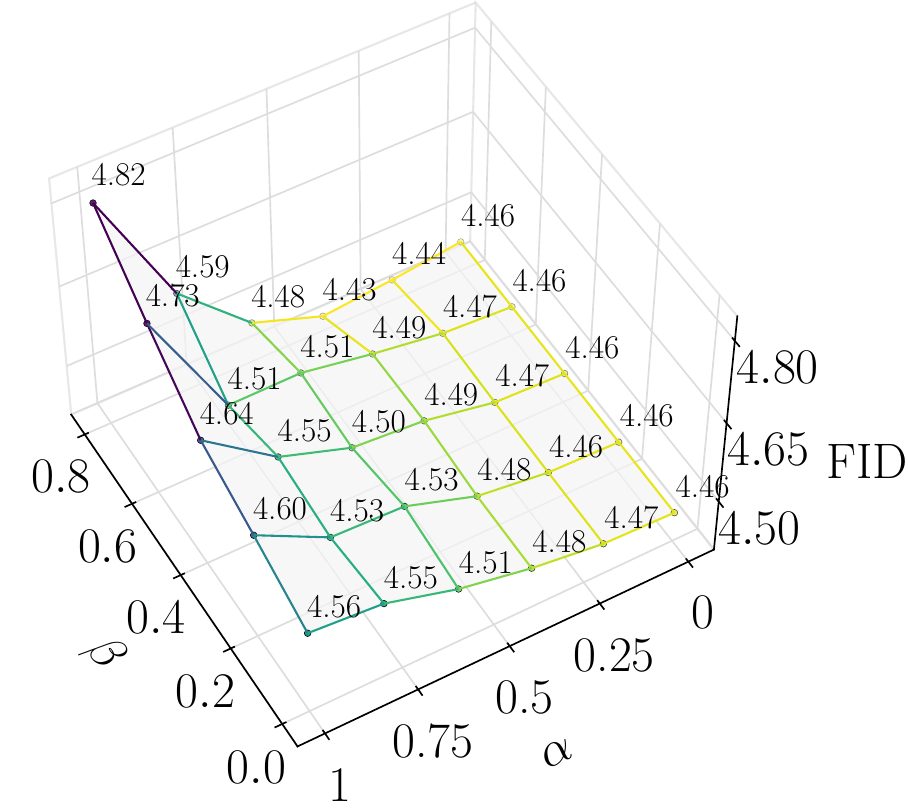}
    \subcaption*{${\textit{NFE}} = 64$}
  \end{subfigure}

  \caption{
FID-10K landscape over MG hyperparameters \((\alpha,\beta)\) at \(\text{CFG}=1.4\).
Moderate MG strengths improve FID over vanilla CFG, especially at lower sampling budgets; the useful region becomes flatter as NFE increases.
}
  \label{fig:3Dsurface-CFG1p4}
\end{figure*}

\begin{figure*}[t]
  \centering

  \begin{subfigure}[t]{0.333\textwidth}
    \centering
    \includegraphics[width=\textwidth]{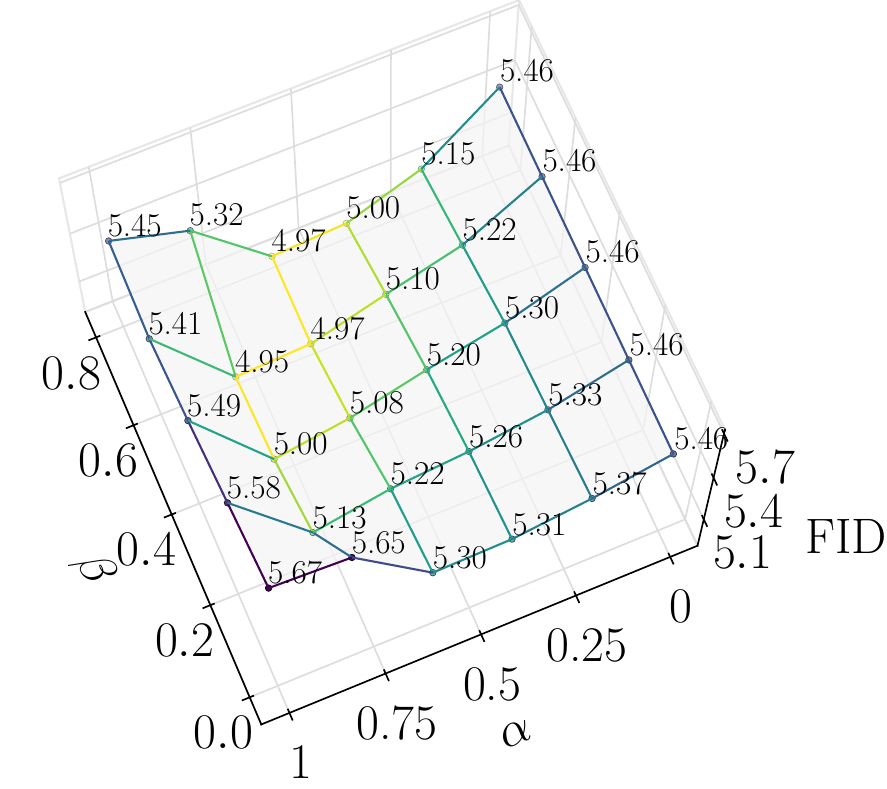}
    \subcaption*{${\textit{NFE}} = 16$}
  \end{subfigure}\hfill
  \begin{subfigure}[t]{0.333\textwidth}
    \centering
    \includegraphics[width=\textwidth]{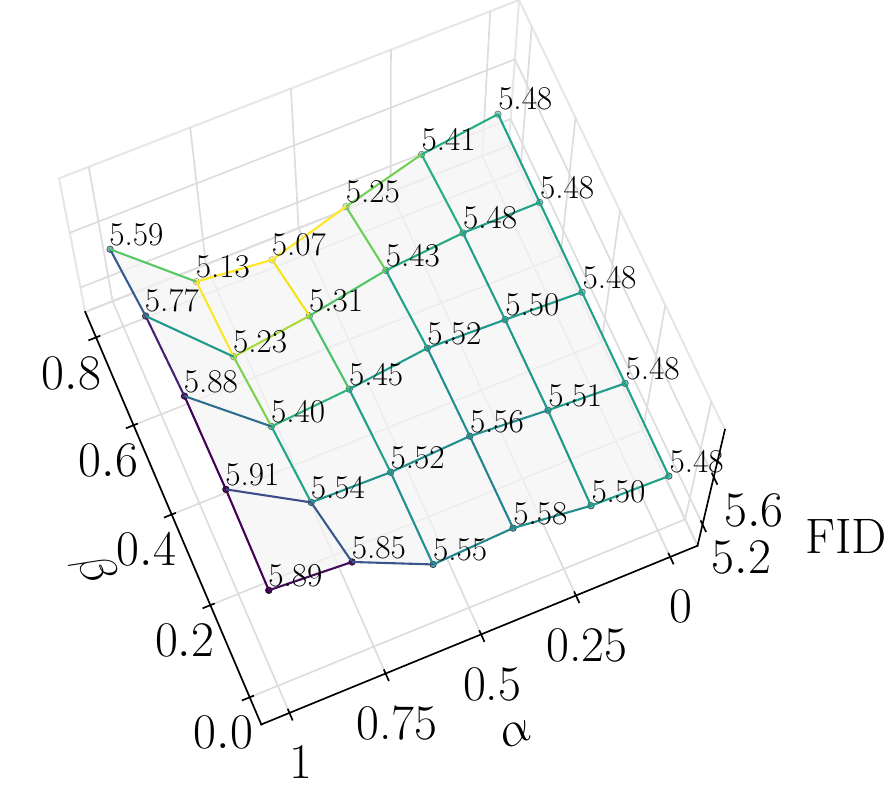}
    \subcaption*{${\textit{NFE}} = 32$}
  \end{subfigure}\hfill
  \begin{subfigure}[t]{0.333\textwidth}
    \centering
    \includegraphics[width=\textwidth]{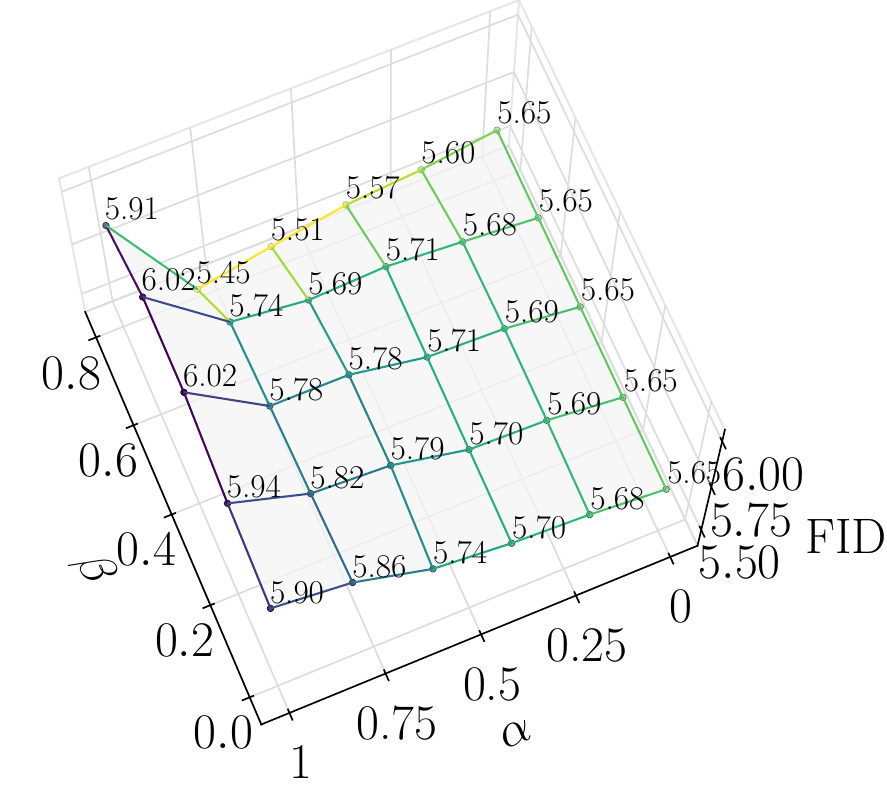}
    \subcaption*{${\textit{NFE}} = 64$}
  \end{subfigure}

  \caption{FID-10K landscape over MG hyperparameters \((\alpha,\beta)\) at \(\text{CFG}=1.6\). Gains remain visible but concentrate in a narrower region, reflecting the stronger sharpening already provided by CFG\@.}
  \label{fig:3Dsurface-CFG1p6}
\end{figure*}

\begin{figure*}[t]
  \centering

  \begin{subfigure}[t]{0.333\textwidth}
    \centering
    \includegraphics[width=\textwidth]{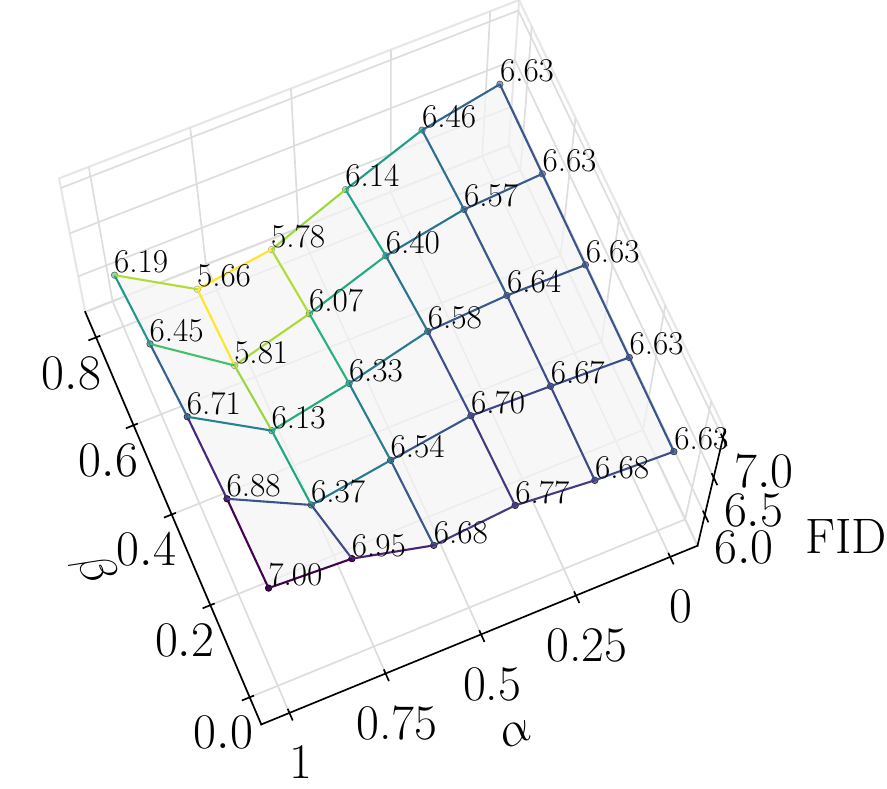}
    \subcaption*{${\textit{NFE}} = 16$}
  \end{subfigure}\hfill
  \begin{subfigure}[t]{0.333\textwidth}
    \centering
    \includegraphics[width=\textwidth]{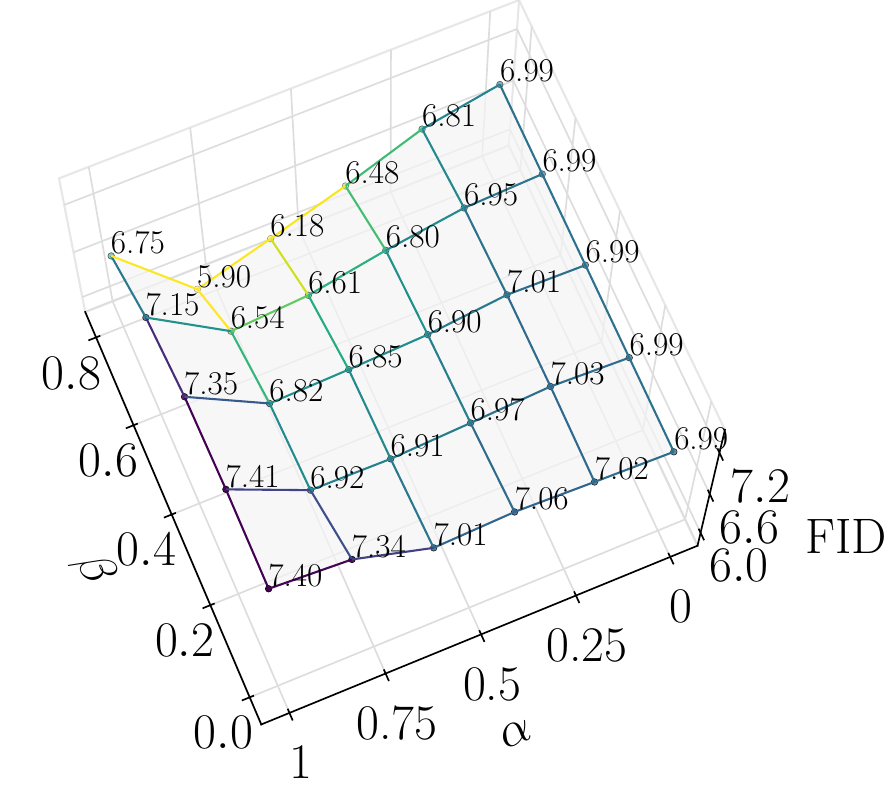}
    \subcaption*{${\textit{NFE}} = 32$}
  \end{subfigure}\hfill
  \begin{subfigure}[t]{0.333\textwidth}
    \centering
    \includegraphics[width=\textwidth]{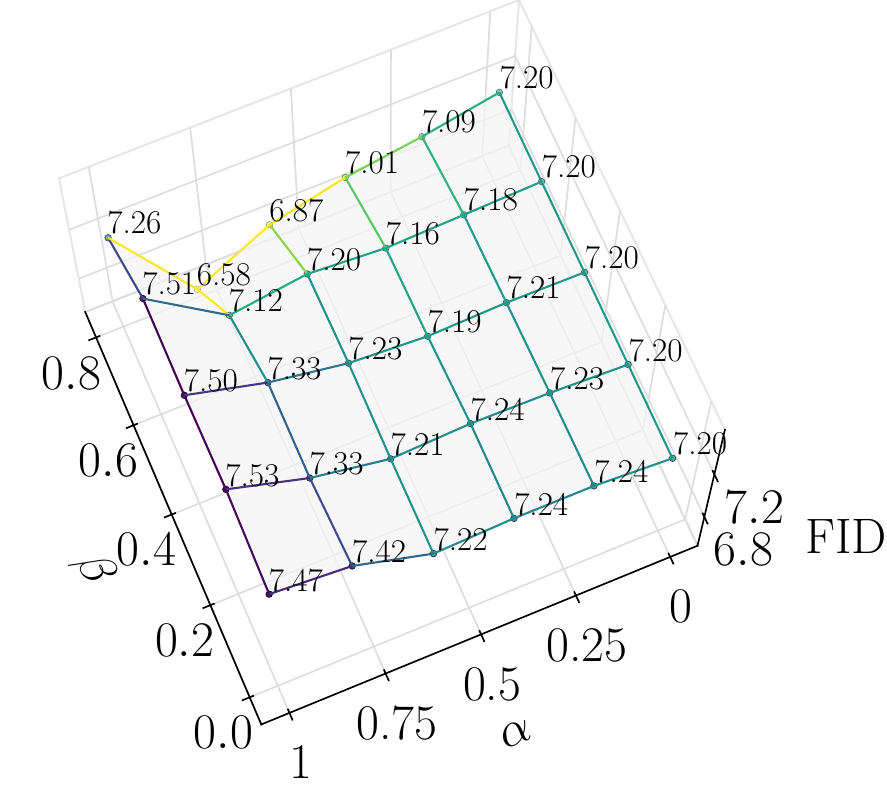}
    \subcaption*{${\textit{NFE}} = 64$}
  \end{subfigure}

  \caption{
FID-10K landscape over MG hyperparameters \((\alpha,\beta)\) at \(\text{CFG}=1.8\).
At this stronger CFG scale, only mild MG strengths are beneficial, while large \(\alpha\) tends to over-correct the trajectory.
}
  \label{fig:3Dsurface-CFG1p8}
\end{figure*}

\begin{figure*}[t]
  \centering

  \begin{subfigure}[t]{0.333\textwidth}
    \centering
    \includegraphics[width=\textwidth]{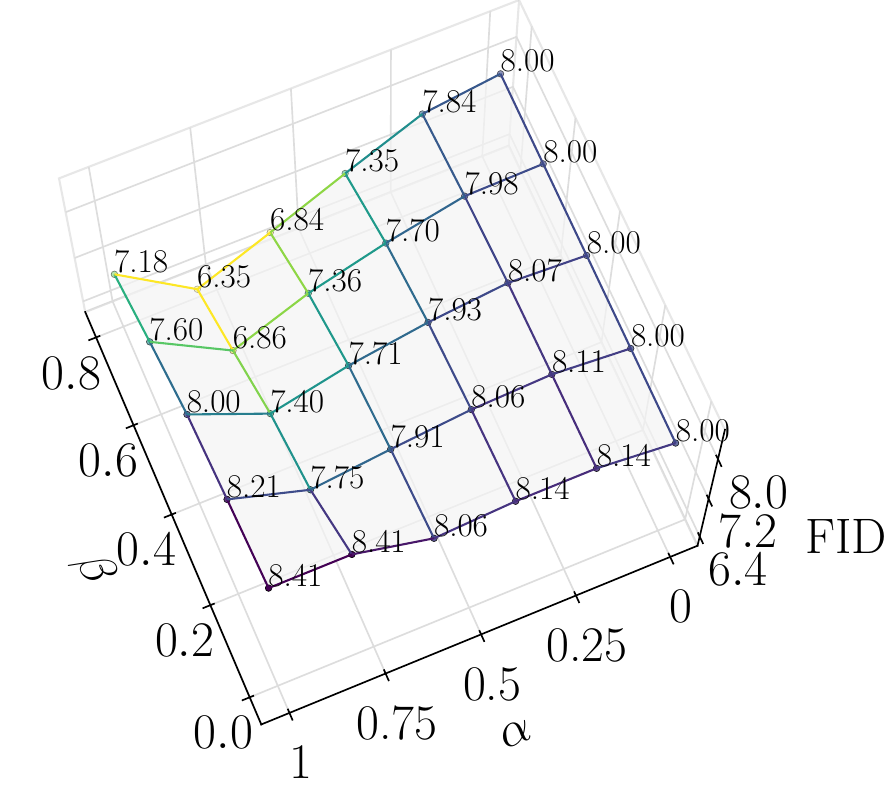}
    \subcaption*{${\textit{NFE}} = 16$}
  \end{subfigure}\hfill
  \begin{subfigure}[t]{0.333\textwidth}
    \centering
    \includegraphics[width=\textwidth]{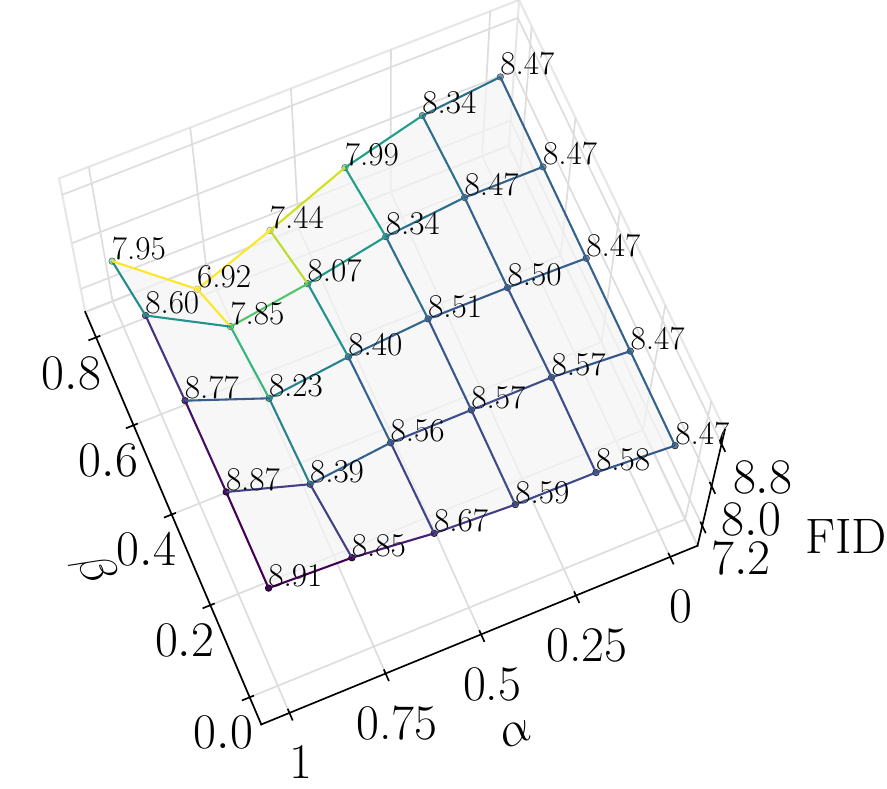}
    \subcaption*{${\textit{NFE}} = 32$}
  \end{subfigure}\hfill
  \begin{subfigure}[t]{0.333\textwidth}
    \centering
    \includegraphics[width=\textwidth]{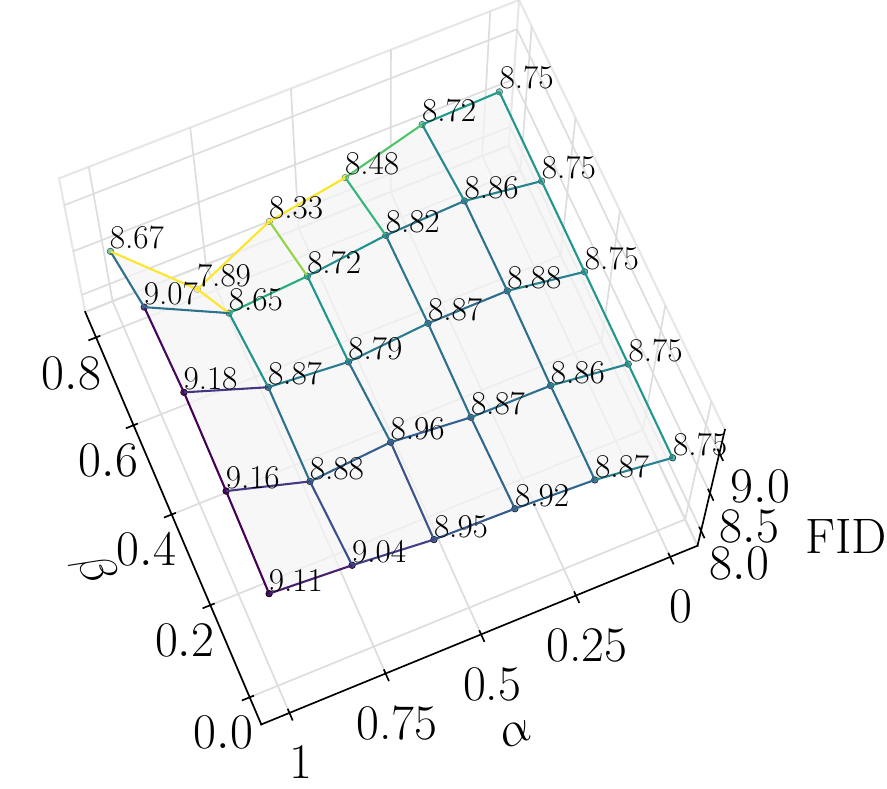}
    \subcaption*{${\textit{NFE}} = 64$}
  \end{subfigure}

  \caption{
FID-10K landscape over MG hyperparameters \((\alpha,\beta)\) at \(\text{CFG}=2.0\).
The surface is dominated by the strong CFG baseline; aggressive MG provides little additional benefit and can degrade FID\@.
}
  \label{fig:3Dsurface-CFG2p0}
\end{figure*}

\paragraph{Additional Ablations on $\alpha$ and $\beta$.}
Figure~\ref{fig:3Dsurface} reports the main FID landscape at \(\text{CFG}=1.2\).
Figures~\ref{fig:3Dsurface-CFG1p0}--\ref{fig:3Dsurface-CFG2p0} extend this sweep to additional guidance strengths and sampling budgets \((\text{NFE}\in\{16,32,64\})\).
Across the no-CFG and moderate-CFG settings, moving away from the \(\alpha=0\) baseline often produces a broad valley of lower FID, especially at 16 and 32 NFEs.
The best regions typically pair moderate \(\alpha\) with small-to-medium \(\beta\), whereas overly large \(\alpha\) or long momentum memory can over-correct the velocity.
As CFG becomes stronger, the useful range of \(\alpha\) narrows, suggesting that MG is most effective when it complements rather than overwhelms the sharpening already induced by CFG\@.

\paragraph{Qualitative Results on ImageNet-256.}
Figures~\ref{fig:imagenet-class248-class278} and~\ref{fig:imagenet-class817-class296} compare standard CFG with MG on ImageNet-256.
All samples use 32 Euler steps and CFG \(=1.5\); MG uses \(\alpha=1.0\), \(\beta=0.6\), and interval \(t\in[0.1,0.7]\).
Across classes, MG reduces structural artifacts and overly smooth local regions while preserving the global composition and the diversity expected from moderate CFG\@.

\begin{figure*}[t]
  \centering
  \includegraphics[width=\linewidth]{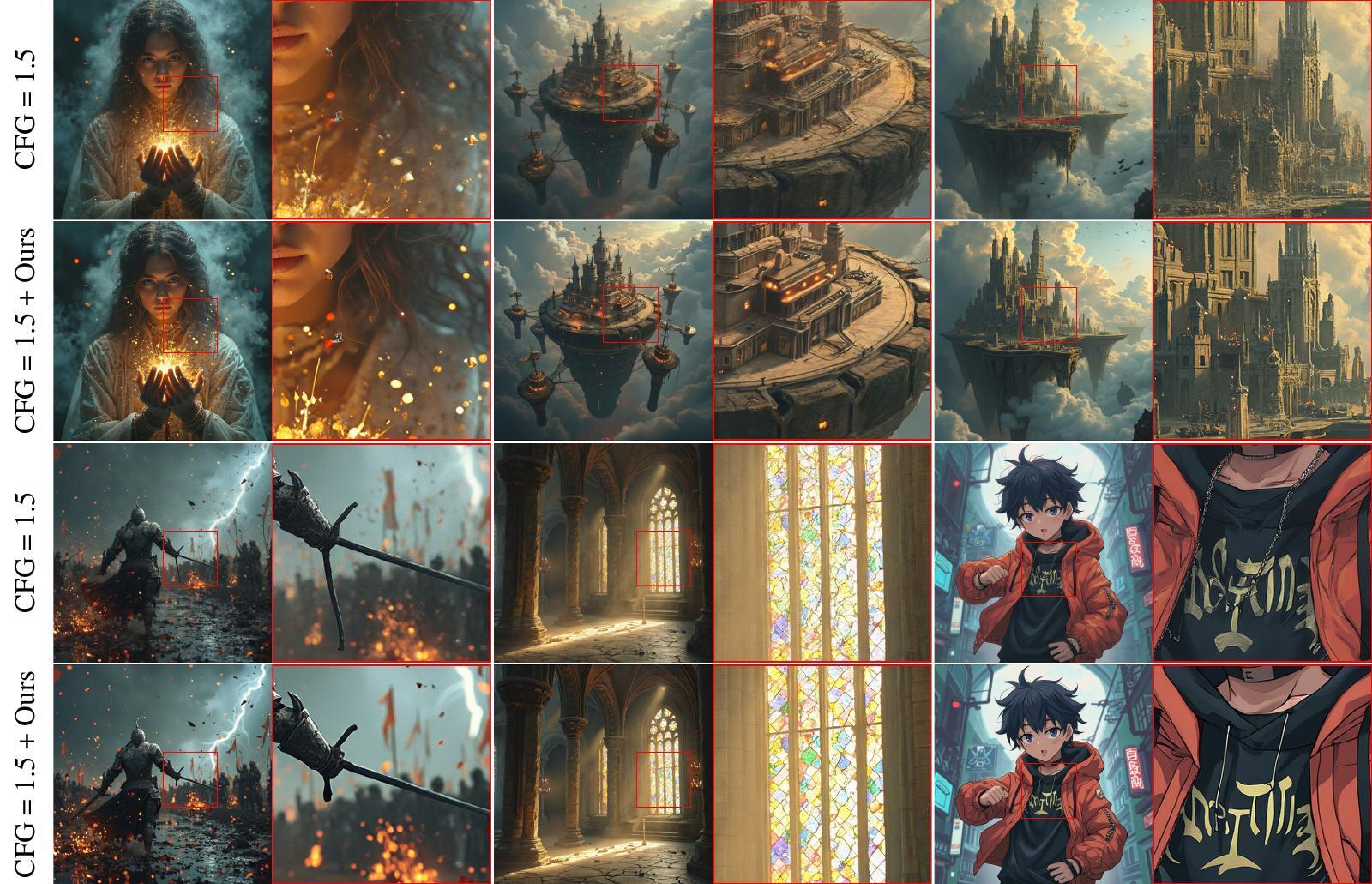}

  \caption{\textbf{Qualitative comparison on FLUX.1-dev at CFG \(=1.5\).} All samples use 50 sampling steps; the second and fourth rows apply MG\@. At this low guidance scale, MG sharpens local textures, separates fine structures more clearly, and improves geometric detail while preserving the overall prompt content.}
  \label{fig:FLUX-CFG1p5cmp}
\end{figure*}

\begin{figure*}[t]
  \centering
  \includegraphics[width=\linewidth]{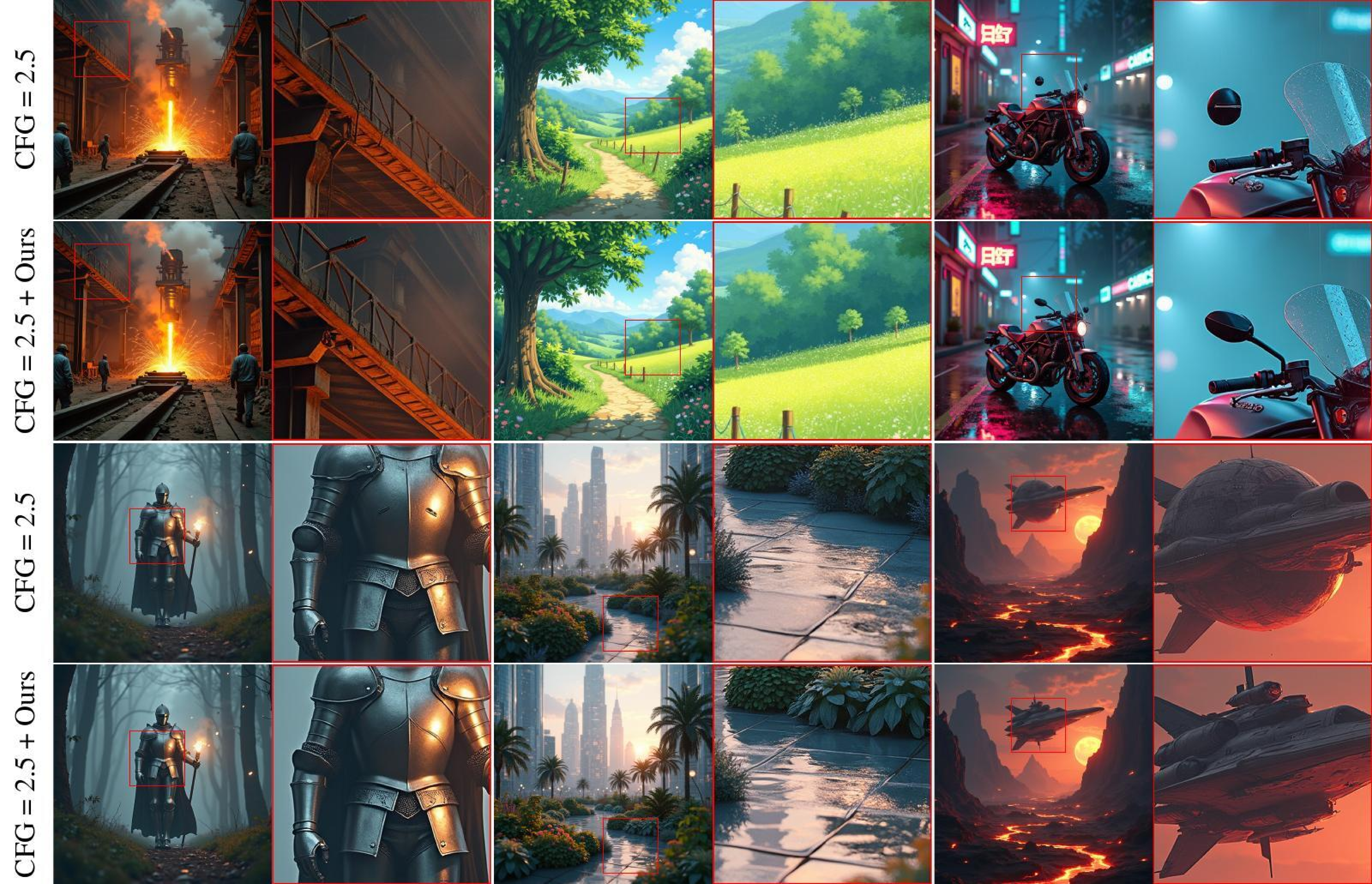}
  \caption{\textbf{Qualitative comparison on FLUX.1-dev at CFG \(=2.5\).} All samples use 50 sampling steps; the second and fourth rows apply MG\@. MG improves structural consistency and local detail, reducing geometry errors and unstable high-frequency textures that appear under CFG alone.}
  \label{fig:FLUX-CFG2p5cmp}
\end{figure*}

\begin{figure*}[t]
  \centering
  \includegraphics[width=\linewidth]{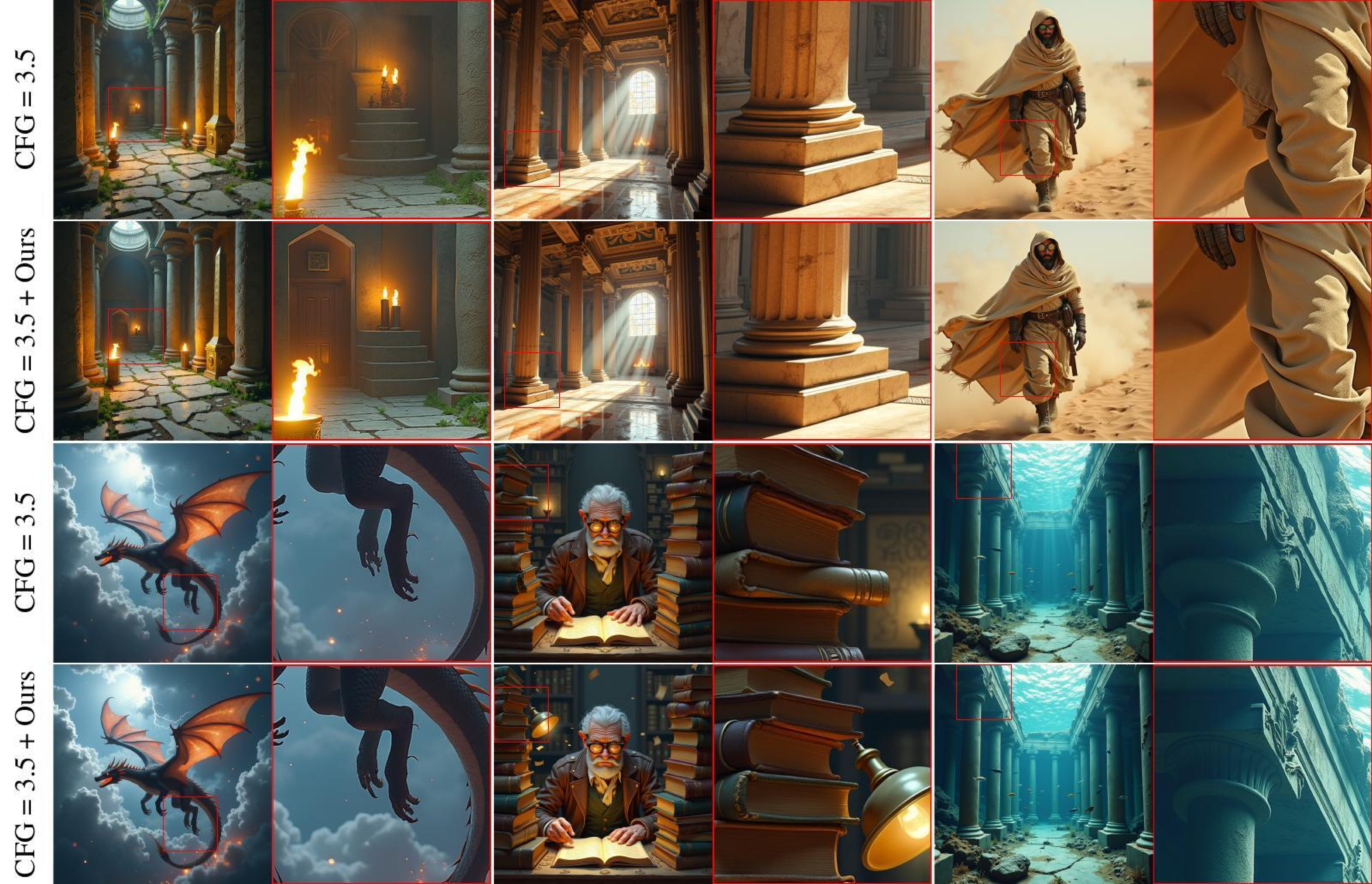}
  \caption{\textbf{Qualitative comparison on FLUX.1-dev at CFG \(=3.5\).} All samples use 50 sampling steps; the second and fourth rows apply MG\@. Under strong CFG, MG helps stabilize fine details, lighting, and object boundaries, reducing oversharpened or distorted structures while retaining the intended scene layout.}
  \label{fig:FLUX-CFG3p5cmp}
\end{figure*}

\paragraph{Qualitative Results on FLUX.1-dev.}
Figures~\ref{fig:FLUX-CFG1p5cmp},~\ref{fig:FLUX-CFG2p5cmp}, and~\ref{fig:FLUX-CFG3p5cmp} compare FLUX.1-dev~\citep{flux2024} samples across three CFG scales.
All samples use 50 sampling steps with the default shifted time discretization.
At low CFG, MG sharpens textures and local structures that are often blurred by the baseline.
At moderate and high CFG, MG mainly stabilizes high-frequency details, shading, and object geometry, reducing the artifacts that can appear when CFG is already strong.

\end{document}